\documentclass{article}

\usepackage{microtype}
\usepackage{graphicx}
\usepackage{subfigure}
\usepackage{booktabs} 

\usepackage{hyperref}

\usepackage{bm} 

\usepackage{listings}

\usepackage{color,soul}

\usepackage{array}  
\usepackage{multirow,bigdelim}

\usepackage{centernot} 
\usepackage[shortlabels]{enumitem} 
\usepackage{amssymb}  

\usepackage{amsthm}  
\newtheorem{definition}{Definition}

\newtheorem{proposition}{Proposition}

\usepackage{float} 
\usepackage{afterpage} 
\usepackage{tikz} 
\usetikzlibrary{positioning}
\usepackage{tikz-dependency}

\usepackage{mathtools} 

\newcommand{\E}{\mathbb{E}}

\usepackage[serbian,english]{babel} 

\newcommand{\independent}{\mathrel{\perp\mspace{-10mu}\perp}}
\newcommand{\dependent}{\centernot{\independent}}

\usepackage[accepted]{icml2019}
\icmltitlerunning{Robustly Disentangled Causal Mechanisms}

\begin{document}

\twocolumn[
\icmltitle{Robustly Disentangled Causal Mechanisms: \\ Validating Deep Representations for Interventional Robustness}



\icmlsetsymbol{equal}{*}

\begin{icmlauthorlist}
\icmlauthor{Raphael Suter}{to}
\icmlauthor{ \DJ or\dj e Miladinovi\'c  }{to}
\icmlauthor{Bernhard Sch\"olkopf}{goo}
\icmlauthor{Stefan Bauer}{goo}
\end{icmlauthorlist}

\icmlaffiliation{to}{Department of Computer Science, ETH Zurich, Switzerland}
\icmlaffiliation{goo}{MPI for Intelligent Systems, T\"ubingen, Germany}

\icmlcorrespondingauthor{Raphael Suter}{rasuter@icloud.com}
\icmlcorrespondingauthor{Stefan Bauer}{bauers@inf.ethz.ch}

\icmlkeywords{Machine Learning, ICML}

\vskip 0.3in
]



\printAffiliationsAndNotice{}  

\begin{abstract}
The ability to learn disentangled representations that split underlying sources of variation in high dimensional, unstructured data is important for data efficient and robust use of neural networks. While various approaches aiming towards this goal have been proposed in recent times, a commonly accepted definition and validation procedure is missing. We provide a causal perspective on representation learning which covers disentanglement and domain shift robustness as special cases. Our causal framework allows us to introduce a new metric for the quantitative evaluation of deep latent variable models. We show how this metric can be estimated from labeled observational data and further provide an efficient estimation algorithm that scales linearly in the dataset size. 
\end{abstract}

\section{Introduction}
Learning deep representations in which different semantic aspects of data are structurally disentangled is of central importance for training robust machine learning models. Separating independent factors of variation could pave the way for successful transfer learning and domain adaptation \citep{bengio2013representation}. Imagine the example of a robot learning multiple tasks by interacting with its environment. For data efficiency, the robot can learn a generic representation architecture that maps its high dimensional sensory data to a collection of general, compact features describing its surrounding. For each task, only a subset of features will be required. If the robot is instructed to grasp an object, it must know the shape and the position of the object, however, its color is irrelevant. On the other hand, when pointing to all red objects is demanded, only position and color are required. 

Having a \emph{disentangled} representation, where each feature captures only one factor of variation, allows the robot to build separate (simple) models for each task based on only a relevant and stable subselection of these generically learned features. We argue that robustness of the learned representation is a crucial property when this is attempted in practice. It has been proposed that features should be selected based on their robustness or invariance across tasks \citep[e.g., ][]{rojas2018causal}, we hence do not want them to be affected by changes in any other factor. In our example, the robot assigned with the grasping task should be able to build a model using features well describing shape and position of the object. For this model to be robust, however, these features must not be affected by changing color (or any other nuisance factor).

It is striking that despite the recent popularity of disentangled representation learning approaches, a commonly accepted definition and validation metric is missing \citep{higgins2018towards}. 
We view disentanglement as a property of a causal process \citep{SpiGlySch93, pearl2009causality} responsible for the data generation, as opposed to only a heuristic characteristic of the encoding. 
Concretely, we call a causal process disentangled when the parents of the generated observations do not affect each other (i.e., there is no total causal effect between them \citep[][Definition 6.12]{peters2017elements}). We call these parents \textit{elementary ingredients}. In the example above, we view color and shape as elementary ingredients, as both can be changed without affecting each other. Still, there can be dependencies between them if for example our experimental setup is confounded by the capabilities of the 3D printers that are used to create the objects (e.g., certain shapes can only be printed in some colors). 

Combining these \emph{disentangled causal processes} with the encoding allows us to study interventional effects on feature representations and estimate them from observational data. 
This is of interest when benchmarking disentanglement approaches based on ground truth data  \citep{locatello2018challenging} or trying to evaluate robustness of a deep representations w.r.t.\ known nuisance factors (e.g., domain changes).
In the example of robotics, knowledge about the generative factors (e.g., the color, shape, weight, etc. of an object to grasp) is often availabe and can be controlled in experiments.

We will start by first giving an overview of previous work in finding disentangled representations and how they have been validated in Section \ref{sec:relatedwork}. In Section \ref{sec:causalmodel} we introduce our framework for the joint treatment of the disentangled causal process and its learned representation. We introduce our notion of interventional effects on encodings and the following \emph{interventional robustness score} in Section \ref{sec:pida} and show how this score can be estimated from \emph{observational} data with an efficient $\mathcal{O}(N)$ algorithm in Section \ref{sec:samplingalgorithm}. Section \ref{sec:experiments} provides experimental evidence in a standard disentanglement benchmark dataset supporting the need of a robustness based disentanglement criterion.

\subsubsection*{Our contributions:} 
\begin{itemize}
\setlength\itemsep{-0.5em}
\item We introduce a unifying causal framework of disentangled generative processes and consequent feature encodings. This perspective allows us to introduce a novel validation metric, the \emph{interventional robustness score}.
\item We show how this metric can be estimated from observational data and provide an efficient algorithm that scales linearly in the dataset size.
\item Our extensive experiments on a standard benchmark dataset show that our robustness based validation is able to discover vulnerabilities of deep representations that have been undetected by existing work.  
\item Motivated by this metric, we additionally present a new visualisation technique which provides an intuitive understanding of dependency structures and robustness of learned encodings.
\end{itemize}
\subsubsection*{Notation:}
We denote the generative factors of high dimensional observations $\bm{X}$ as $\bm{G}$. The latent variables learned by a model, e.g., a variational auto-encoder (VAE) \citep{kingma2013auto}, are denoted as $\bm{Z}$. We use the notation $E(\cdot)$ to describe the encoding which in case of VAEs corresponds to the posterior mean of $q_\phi(\bm{z}|\bm{x})$.  Capital letters denote random variables, and lower case observations thereof. Subindices $\bm{Z}_J$ for a set $J$ or $Z_j$ for a single index $j$ denote the selected components of a multidimensional variable. A backslash $\bm{Z}_{\backslash J}$ denotes all components except those in $J$.

\section{Related Work}\label{sec:relatedwork}
In the framework of variational auto-encoders (VAEs) \citep{kingma2013auto} the (high dimensional) observations $\bm{x}$ are modelled to be generated from some latent features $\bm{z}$ with chosen prior $p(\bm{z})$ according to the probabilistic model $p_\theta(\bm{x}|\bm{z})p(\bm{z})$. The generative model $p_\theta(\bm{x}|\bm{z})$ as well as the proxy posterior $q_\phi(\bm{z}|\bm{x})$ can be estimated using neural networks by maximizing the variational lower bound (ELBO) of $\log p(\bm{x}_1,\dots,\bm{x}_N)$:
\begin{equation}\label{eq:vaeelbo}
\resizebox{.99\hsize}{!}{$
\mathcal{L}_{VAE} = \sum_{i=1}^N  \E_{q_\phi(\bm{z}|\bm{x}^{(i)})} [ \log p_\theta (\bm{x}^{(i)} | \bm{z}) ] 
- D_{KL}( q_\phi(\bm{z}|\bm{x}^{(i)}) \| p (\bm{z}) )$}. 
\end{equation}
This objective function a priori does not encourage much structure on the latent space (except some similarity to the chosen prior $p (\bm{z})$ which is usually isotropic Gaussian). More precisely, for a given encoder $E$ and decoder $D$ any bijective transformation $g$ of the latent space $\bm{z}=E(\bm{x})$ yields the same reconstruction $\hat{\bm{x}} = D(g(g^{-1}(E(\bm{x}))) = D(E(\bm{x}))$.

Various proposals for more structure imposing regularization have been made, either with some sort of supervision \citep[e.g.][]{siddharth2017learning, bouchacourt2017multi, liu2017detach, mathieu2016disentangling, cheung2014discovering} or completely unsupervised \citep[e.g.][]{higgins2016beta, kim2018disentangling, chen2018isolating, kumar2017variational, esmaeili2018hierarchical}. 
\citet{higgins2016beta} proposed the $\beta$-VAE penalizing the Kullback-Leibler divergence (KL) term in the VAE objective \eqref{eq:vaeelbo} more strongly, which encourages similarity to the factorized prior distribution. Others used techniques to encourage statistical independence between the different components in $\bm{Z}$, e.g., FactorVAE \citep{kim2018disentangling} or $\beta$-TCVAE \citep{chen2018isolating}, similar to independent component analysis \citep[e.g.][]{comon1994independent}. With disentangling the \textit{inferred prior} (DIP-VAE), \citet{kumar2017variational} proposed encouraging factorization of $q_\phi(\bm{z}) = \int q_\phi (\bm{z} | \bm{x}) p(\bm{x}) \, d\bm{x}$.

A special form of structure in the latent space which has gained a lot of attention in recent time is referred to as \emph{disentanglement} \citep{bengio2013representation}. This term encompasses the understanding that each learned feature in $\bm{Z}$ should represent structurally different aspects of the observed phenomena (i.e., capture different sources of variation).

Various methods to validate a learned representation for disentanglement based on known ground truth generative factors $\bm{G}$ have been proposed \citep[e.g.][]{eastwood2018framework, ridgeway2018learning, chen2018isolating, kim2018disentangling}. While a universal definition of disentanglement is missing, the most widely accepted notion is that one feature $Z_i$ should capture information of only one generative factor \citep{eastwood2018framework, ridgeway2018learning}. This has for example been expressed as the mutual information of a single latent dimension $Z_i$ with generative factors $G_1,\dots,G_K$ \citep{ridgeway2018learning}, where in the ideal case each $Z_i$ has some mutual information with one generative factor $G_k$ but none with all the others. Similarly, \citet{eastwood2018framework} trained predictors (e.g., Lasso or random forests) for a generative factor $G_k$ based on the representation $\bm{Z}$. In a disentangled model, each dimension $Z_i$ is only useful (i.e., has high feature importance) to predict one of those factors (see appendix \ref{sec:experimentaldetails} for details).

Validation without known generative factors is still an open research question and so far it is not possible to quantitatively validate disentanglement in an unsupervised way. The community has been using "latent traversals" (i.e., changing one latent dimension and subsequently re-generating the image) for visual inspection when supervision is not available \citep[see e.g.][]{chen2018isolating}. This can be used to encounter physically meaningful interpretations of each dimension.

\section{Causal Model}\label{sec:causalmodel}

We will first consider assumptions for the causal process underlying the data generating mechanism. Following this, we discuss consequences for trying to match encodings $\bm{Z}$ with causal factors $\bm{G}$ in a deep latent variable model.

\subsection{Disentangled Causal Model}
As opposed to previous approaches that defined disentanglement heuristically as properties of the learned latent space, we take a step back and first introduce a notion of disentanglement on the level of the true causal mechanism (or data generation process). Subsequently, we can use this definition to better understand a learned probabilistic model for latent representations and evaluate its properties.

We assume to be given a set of observations from a (potentially high dimensional) random variable $\bm{X}$. 
In our model, the data generating process is described by $K$ causes of variation (generative factors) $\bm{G} = [G_1,\dots,G_K]$ (i.e., $\bm{G} \rightarrow \bm{X}$) that do not cause each other. These factors $\bm{G}$ are generally assumed to be unobserved and are objects of interest when doing deep representation learning. In particular, knowledge about $\bm{G}$ could be used to build lower dimensional predictive models, not relying on the (unstructured) $\bm{X}$ itself. This could be classic prediction of a label $Y$, often in "confounded" direction (i.e., predicting effects from other effects) if $\bm{G} \rightarrow (\bm{X}, Y)$ or in anti-causal direction if $Y \rightarrow \bm{G} \rightarrow \bm{X}$. It is also relevant in a domain change setting when we know that the domain $S$ has an impact on $\bm{X}$, i.e., $(S, \bm{G})  \rightarrow \bm{X}$.

Having these potential use cases in mind, we assume the generative factors themselves to be confounded by (multi-dimensional) $\bm{C}$, which can for example include a potential label $Y$ or source $S$. Hence, the resulting causal model $\bm{C} \rightarrow \bm{G} \rightarrow \bm{X}$ allows for statistical dependencies between latent variables $G_i$ and $G_j$, $i \neq j$, when they are both affected by a certain label, i.e., $G_i \leftarrow Y \rightarrow G_j$.

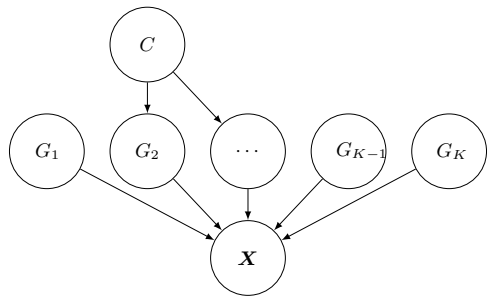
\begin{figure}[!htb]
            \centering
\resizebox{1.4\totalheight}{!}{
            \begin{tikzpicture}[
            node distance=0.5cm and 0.4 cm,
            mynode/.style={draw,circle, align=center, inner sep=0pt, text width=4mm, minimum size = 1.2cm},
            ]
            \node[mynode](g1){\small $G_1$};
            \node[mynode, right =of g1] (g2){\small $G_2$};
            \node[mynode, right =of g2](dots){\small $\cdots$};
            \node[mynode, right =of dots](gkm1){\small $G_{K-1}$};
            \node[mynode, right =of gkm1](gk){\small $G_K$};
            \node[mynode, above  =of g2](c){\small $C$};
            \node[mynode, below  =of dots] (x) {\small $\bm{X}$};
            \path
            (c) edge[-latex] (g2)
            (c) edge[-latex] (dots)
            (g1) edge[-latex] (x)
            (g2) edge[-latex] (x)
            (dots) edge[-latex] (x)
            (gkm1) edge[-latex] (x)
            (gk) edge[-latex] (x)
            ;
            \end{tikzpicture}
            }
            \caption{\textbf{Disentangled Causal Mechanism:} This graphical model encompasses our assumptions on a disentangled causal model. $C$ stands for a confounder, $\bm{G} = (G_1, G_2, \dots, G_K)$ are the generative factors (or elementary ingredients) and $\bm{X}$ the observed quantity. In general, there can be multiple confounders affecting a range of elementary ingredients each.}
            \label{fig:graphicalmodel}
\end{figure}

However, a crucial assumption of our model is that these latent factors should represent \textit{elementary ingredients} to the causal mechanism generating $\bm{X}$ (to be defined below), which can be thought of as descriptive features of $\bm{X}$ that can be changed without affecting each other (i.e., there is no causal effect between them). A similar assumption on the underlying model is likewise a key requirement for the recent extension of identificability results of non-linear ICA \citep{hyvarinen2018nonlinear}. We formulate this assumption of a disentangled causal model as follows (see also Figure \ref{fig:graphicalmodel}):
\begin{definition}[Disentangled Causal Process]\label{def:disentangledmodel}
Consider a causal model for $\bm{X}$ with generative factors $\bm{G}$, described by the mechanisms 
$p(\bm{x} | \bm{g})$, 
where $\bm{G}$ could generally be influenced by $L$ confounders $\bm{C}=(C_1,\dots,C_L)$. This causal model for $\bm{X}$ is called disentangled if and only if it can be described by a structural causal model (SCM) \citep{pearl2009causality} of the form
\resizebox{.99\linewidth}{!}{
  \begin{minipage}{\linewidth}
 \begin{align*}
\bm{C} \leftarrow & \bm{N}_c \\
G_i \leftarrow & f_i(\bm{PA}^C_i, N_i), \quad   \bm{PA}^C_i \subset \{C_1, \dots, C_L \}, ~ i = 1,\dots,K \\
\bm{X} \leftarrow & g(\bm{G}, N_x)
\end{align*}
  \end{minipage}
}
with functions $f_i, g$ and jointly independent noise variables $\bm{N}_c, N_1,\dots,N_K, N_x$. Note that $\forall i \neq j$ $G_i \not\rightarrow G_j $.
\end{definition}
In practice we assume that the dimensionality of the confounding $L$ is significantly smaller than the number of factors $K$.
%

%

This definition reflects our understanding of elementary ingredients $G_i, i=1,\dots,K$, of the causal process. Each ingredient should work on its own and is changable without affecting others. This reflects the {\em independent mechanisms (IM)} assumption \citep{scholkopf2012causal}. Independent mechanisms as components of causal models allow intervention on one mechanism without affecting the other modules and thus correspond to the notion of  \textit{independently controllable factors} in reinforcement learning \citep{thomas2017independently}. Our setting is broader, describing any causal process and inheriting the generality of the notion of IM, pertaining to autonomy, invariance and modularity \citep{peters2017elements}.

Based on this view of the data generation process, we can prove (see Appendix \ref{app:proof}) the following observations which will help us discuss notions of disentanglement and deep latent variable models. 

\begin{proposition}[Properties of a Disentangled Causal Process]\label{prop:properties}
A disentangled causal process as introduced in Definition \ref{def:disentangledmodel} fulfills the following properties:
\begin{enumerate}[(a)]
\setlength\itemsep{-0.5em}
\item \label{item:causalmechanism} $p(\bm{x}|\bm{g})$ describes a causal mechanism invariant to changes in the distributions $p(g_i)$. 
\item \label{item:dependentz} In general, the latent causes can be dependent 
\begin{equation*}
G_i \dependent G_j, i \neq j.
\end{equation*}
Only if we condition on the confounders in the data generation they are independent 
\begin{equation*}
G_i \independent G_j | \bm{C} \quad \forall i \neq j.
\end{equation*}
\item \label{item:dependentzgivenx} Knowing what observation of $\bm{X}$ we obtained renders the different latent causes dependent, i.e., 
\begin{equation*}
G_i \dependent G_j | \bm{X}.
\end{equation*}
\item \label{item:processinginequality} The latent factors $\bm{G}$ already contain all information about confounders $\bm{C}$ that is relevant for $\bm{X}$, i.e.,
$$I(\bm{X}; \bm{G}) = I(\bm{X}; (\bm{G}, C)) \geq I(\bm{X}; C)$$ where $I$ denotes the mutual information.
\item \label{item:interventiononz} There is no total causal effect from $G_j$ to $G_i$ for $j \neq i$; i.e., intervening on $G_j$ does not change $G_i$, i.e,
\begin{equation*}
\forall g_j^{\triangle} \quad p(g_i | \text{do}(G_j \leftarrow g_j^{\triangle})) = p(g_i) \quad \left( \neq p(g_i | g_j^{\triangle}) \right)
\end{equation*}
\item \label{item:adjustmentset} The remaining components of $\bm{G}$, i.e., $\bm{G}_{\backslash j}$, are a valid adjustment set \citep{pearl2009causality} to estimate interventional effects from $G_j$ to $\bm{X}$ based on observational data, i.e.,
\begin{equation*}
p(\bm{x} | \text{do}(G_j \leftarrow g_j^{\triangle})) = \int p(\bm{x} | g_j^{\triangle}, \bm{g}_{\backslash j}) p(\bm{g}_{\backslash j}) \, d \bm{g}_{\backslash j}.
\end{equation*}
\item \label{item:noconfounding} If there is no confounding, conditioning is sufficient to obtain the post interventional distribution of $\bm{X}$:
\begin{equation*}
p(\bm{x} | \text{do}(G_j \leftarrow g_j^{\triangle}))  = p(\bm{x} |  g_j^{\triangle})
\end{equation*}
\end{enumerate}
\end{proposition}

\subsection{Disentangled Latent Variable Model}
We can now understand generative models with latent variables (e.g., the decoder $p_\theta (\bm{x} | \bm{z})$ in VAEs) as models for the causal mechanism in \ref{item:causalmechanism} and the inferred latent space through $q_\phi (\bm{z} | \bm{x})$ as proxy to the generative factors $\bm{G}$. Property \ref{item:processinginequality} gives hope that under an adequate information bottleneck we can indeed recover information about causal parents and not the confounders.
Ideally, we would hope for a one-to-one correspondance of $Z_i$ to $G_i$ for all $i=1,\dots,K$. In some situations it might be useful to learn multiple latent dimensions for one causal factor for a more natural description, e.g., describing an angle $\theta$ as $\cos(\theta)$ and $\sin(\theta)$ \citep{ridgeway2018learning}. Hence, we will generally allow the encodings $\bm{Z}$ to be $K^\prime$ dimensional, where usually $K^\prime \geq K$. The
$\beta$-VAE \citep{higgins2016beta} encourages factorization of $q_\phi(\bm{z}|\bm{x})$ through penalization of the KL to its prior $p(\bm{z})$. Due to property \ref{item:dependentzgivenx} other approaches were introduced making use of statistical independence \citep{kim2018disentangling, chen2018isolating, kumar2017variational}.
\citet{esmaeili2018hierarchical} allow dependence within groups of variables in a hierarchical model (i.e., with some form of confounding where property \ref{item:dependentz} becomes an issue) by specifically modelling groups of dependent latent encodings. In contrast to the above mentioned approaches, this requires prior knowledge on the generative structure.
We will make use of property \ref{item:adjustmentset} to solve the task of using observational data to evaluate deep latent variable models for disentanglement and robustness.
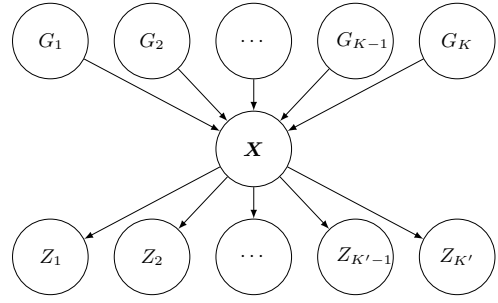
\begin{figure}[!tb]
            \centering
\resizebox{1.4\totalheight}{!}{
 \begin{tikzpicture}[
            node distance=0.5 cm and 0.4 cm,
            mynode/.style={draw,circle, align=center, inner sep=0pt, text width=6mm, minimum size = 1.2cm},
            ]
            \node[mynode](g1){\small $G_1$};
            \node[mynode, right =of g1] (g2){\small $G_2$};
            \node[mynode, right =of g2](dots){\small $\cdots$};
            \node[mynode, right =of dots](gkm1){\small $G_{K-1}$};
            \node[mynode, right =of gkm1](gk){\small $G_K$};
            \node[mynode, below  =of dots] (x) {\small $\bm{X}$};
            \node[mynode, below =of x](zdots){\small $\cdots$};
            \node[mynode, left =of zdots] (z2){\small $Z_2$};
            \node[mynode, left = of z2](z1){\small $Z_1$};
            \node[mynode, right =of zdots](zkm1){\small $Z_{K^\prime-1}$};
            \node[mynode, right =of zkm1](zk){\small $Z_{K^\prime}$};
            \path
            (g1) edge[-latex] (x)
            (g2) edge[-latex] (x)
            (dots) edge[-latex] (x)
            (gkm1) edge[-latex] (x)
            (gk) edge[-latex] (x)
            (x) edge[-latex] (z1)
            (x) edge[-latex] (z2)
            (x) edge[-latex] (zdots)
            (x) edge[-latex] (zkm1)
            (x) edge[-latex] (zk)
            ;
            \end{tikzpicture}}
            \caption{We assume that the data are generated by a process involving a set of unknown independent mechanisms $G_i$ (which may themselves be confounded by other processes, see Figure~\ref{fig:graphicalmodel}). In the simplest case, disentangled representation learning aims to recover variables $Z_i$ that capture the independent mechanisms $G_i$ in the sense that they (i) represent the information contained in the $G_i$ and (ii) respect the causal generative structure of $\bm{G}\to \bm{X}$ in an interventional sense: in particular, for any $i$, localized interventions on another cause $G_j$ $(j\neq i)$ should not affect $Z_i$. In practice, there need not be a direct correspondence between $G_i$ and $Z_i$ variables (e.g., multiple latent variables may jointly represent one cause), hence our definitions deal with sets of factors rather than individual ones.
Note that in the unsupervised setting, we do not know $\bm{G}$ nor the mapping from $\bm{G}$ to $\bm{X}$ (we do know, however, the ``decoder'' mapping from $\bm{Z}$ to $\bm{X}$, not shown in this picture). In experimental evaluations of disentanglement, however, such knowledge is usually assumed.}
            \label{fig:unifiedframework}
\end{figure}
Figure \ref{fig:unifiedframework} illustrates our causal perspective on representation learning which encompasses the data generating process ($\bm{G} \rightarrow \bm{X}$) as well as the subsequent encoding through $E(\cdot)$ ($\bm{X} \rightarrow \bm{Z}$). Based on this viewpoint, we define the interventional effect of a \emph{group} of generative factors $\bm{G}_J$ on the implied latent space encodings $\bm{Z}_{L}$ with proxy posterior $q_\phi  (\bm{z} |\bm{x})$ from a VAE, where $J \subset \{1,\dots,K \}$ and $L \subset \{1,\dots,K^\prime \}$ as: \begin{equation*}
\label{eq:interventiononz}
\resizebox{.99\hsize}{!}{$
p(\bm{z}_{I} | \text{do}(\bm{G}_J \leftarrow \bm{g}_J^{\triangle}) ) := \int q_\phi (\bm{z}_I |\bm{x}) \, p(\bm{x}| \text{do}(\bm{G}_J \leftarrow \bm{g}_J^{\triangle})) \, d\bm{x}$ }
\end{equation*}
This definition is consistent with the above graphical model as it implies that $p(\bm{z}_{I} | \bm{x}, \text{do}(\bm{G}_J \leftarrow \bm{g}_J^{\triangle}) ) = q_\phi(\bm{z}_{I} | \bm{x} )$.

\section{Interventional Robustness}\label{sec:pida}
Building on the definition of interventional effects on deep feature representations in Eq.~\eqref{eq:interventiononz}, we now derive a robustness measure of encodings with respect to changes in certain generative factors. 

Let $L \subset \{1,\dots,K^\prime \}$ and $I,J \subset \{1,\dots,K \}, I \cap J = \emptyset$ be groups of indices in the latent space and generative space. For generality, we will henceforth talk about robustness of \emph{groups} of features $\bm{Z}_L$ with respect to interventions on \emph{groups} of generative factors $\bm{G}_J$. We believe that having this general formulation of allowing disagreements between groups of latent dimensions and generative factors provides more flexibility, for example when multiple latent dimensions are used to describe one phenomenon \citep{esmaeili2018hierarchical} or when some sort of supervision is available through groupings in the dataset according to generative factors \citep{bouchacourt2017multi}. Below, we will also discuss special cases of how these sets can be chosen.

If we assume that the encoding $\bm{Z}_L$ captures information about the causal factors $\bm{G}_I$ and we would like to build a predictive model that only depends on those factors, we might be interested in knowing how robust our encoding is with respect to nuisance factors $\bm{G}_J$, where $I \cap J = \emptyset$. To quantify this robustness for specific realizations of $\bm{g}_I$ and $\bm{g}_J^{\triangle}$ we make the following definition:
\begin{definition}[Post Interventional Disagreement]
\label{def:pida}
For any given set of feature indices $L \subset \{1,\dots,K^\prime \}$, $\bm{g}_I$ and $\bm{g}_J^{\triangle}$, we call
\begin{align*}\label{eq:pida}
&\mathrm{PIDA}(L | \bm{g}_I, \bm{g}_J^{\triangle}) :=  \\  \nonumber  & d \left( \E [\bm{Z}_L | \text{do}(\bm{G}_I \leftarrow \bm{g}_I)], \, \E [\bm{Z}_L | \text{do}(\bm{G}_I \leftarrow \bm{g}_I, \bm{G}_J \leftarrow \bm{g}_J^{\triangle})] \right)&
\end{align*}
the post interventional disagreement ($\mathrm{PIDA}$) in $\bm{Z}_L$ due to $\bm{g}_J^{\triangle}$ given $\bm{g}_I$. Here, $d$ is a suitable distance function (e.g., $\ell_2$-norm).
\end{definition}
The above definition on its own is likewise a contribution to the defined but unused notion of  extrinsic disentanglement in \citet{besserve2018counterfactuals}. 
$\mathrm{PIDA}$ now quantifies the shifts in our inferred features $\bm{Z}_L$ we experience when the generative factors $\bm{G}_J$ are externally changed to $\bm{g}_J^{\triangle}$ while the generative factors that we are actually interested in capturing with $\bm{Z}_L$ (i.e., $\bm{G}_I$) remain at the predefined setting of $\bm{g}_I$. Using expected values after \emph{intervention} on the generative factors (i.e., Pearl's do-notation), as opposed to regular conditioning, allows for interpretation of the score also when factors are dependent due to confounding. The do-notation represents setting these generative values by external intervention. It thus isolates the {\em causal} effect that a generative factor has, which in general is not possible using standard conditioning \citep{pearl2009causality}.
This neglects the history that might have led to the observations in the collection phase of the \emph{observational} dataset.
For example, when a robot is trained with various objects of different colors, it might be the case that certain shapes occur more often in specific colors (e.g., due to 3D printer capabilities). When we would condition the feature encoding on on a specific color, the observed effects might as well be due to a change in object shape. The interventional distribution, on the other hand, measures by definition the change features experience due to externally setting the color while all other generative factors remain the same.
If there was no confounding in the generative process, this definition is equivalent to regular conditioning (see Proposition \ref{prop:properties} \ref{item:noconfounding}). 

For robustness reasons, we are interested in the worst case effect any change in nuisance parameters $\bm{g}_J^{\triangle}$ might have. We call this the maximal post interventional disagreement ($\mathrm{MPIDA}$): $
\mathrm{MPIDA}(L | \bm{g}_I, J) := \sup_{\bm{g}_J^{\triangle}}  \mathrm{PIDA}(L | \bm{g}_I, \bm{g}_J^{\triangle})$.
This metric is still computed for a specific realization of $\bm{G}_I$. Hence, we weight this score according to occurance probabilities of $\bm{g}_I$, which leads us to the expected $\mathrm{MPIDA}$:
$\mathrm{EMPIDA}(L | I, J) := \E_{\bm{g}_I} \left[ \mathrm{MPIDA}(L | \bm{g}_I, J) \right]$.
$\mathrm{EMPIDA}$ is now a (unnormalized) measure in $[0, \infty)$ quantifying the worst-case shifts in the inferred $\bm{Z}_L$ we have to expect due to changes in $\bm{G}_J$ even though our generative factors of interest $\bm{G}_I$ remain the same.
This is for example of interest when the robot in our introductory example learns a generic feature representation $\bm{Z}$ of his environment from which he wants to make a subselection of features $\bm{Z}_L$ in order to perform a grasping task. For this model to work well, the generative factor of the object $I = \{ \texttt{shape}, \texttt{weight} \}$ are important, however, factor $J = \{ \texttt{color} \}$ is not. Now, the robot can evaluate how robust its features $\bm{Z}_L$ perform at the task requiring $I$ but not $J$.

We propose to normalize this quantity with $\mathrm{EMPIDA}(L | \emptyset, \{ 1,\dots,K\})$, which represents the expected maximal deviation from the mean encoding of $\bm{Z}_L$ without fixed generative factors as it is often useful to have a normalized score for comparisons. Hence, we define:
\begin{definition}[Interventional Robustness Score]
\begin{equation}\label{eq:irs}
\mathrm{IRS}(L| I, J) := 1 - \frac{\mathrm{EMPIDA}(L | I, J)}{\mathrm{EMPIDA}(L | \emptyset, \{ 1,\dots,K\})}
\end{equation}
\end{definition}
This score yields $1.0$ for perfect robustness (i.e., no harm is done by changes in $\bm{G}_J$) and $0.0$ for no robustness. Note that $\mathrm{IRS}$ has a similar interpretion to a $R^2$ value in regression. Instead of measuring the captured variance, it looks at worst case deviations of inferred values.

\paragraph{Special Case: Disentanglement}
One important special case includes the setting where $L=\{l \}$, $I = \{ i \}$ and $J = \{1,\dots, i-1,i+1,\dots,K\}$. This corresponds to the degree to which $Z_l$ is robustly isolated from any extraneous causes (assuming $Z_l$ captures $G_i$), which can be interpreted as the concept of \textit{disentanglement} in the framework of \citet{eastwood2018framework}. 
We define 
\begin{equation}\label{eq:disentanglementscore}
D_l := \max_{i \in \{1,\dots,K \}} \mathrm{IRS}(\{l \} | \{ i \}, \{1,\dots,K \} \backslash \{ i \})
\end{equation}
as disentanglement score of $Z_l$. The maximizing $i^\star$ is interpreted as the generative factor that $Z_l$ captures predominantly. Intuitively, we have robust disentanglement when a feature $Z_l$ reliably captures information about the generative factor $G_{i^\star}$, where reliable means that the inferred value is always the same when $g_{i^\star}$ stays the same, regardless of what the other generative factors $\bm{G}_{\backslash i^\star}$ are doing.

In our evaluations of disentanglement, we also plot the full dependency matrix $\hat{\bm{R}}$ with $\hat{R}_{li} = \mathrm{IRS}(\{l \} | \{ i \}, \{1,\dots,K \} \backslash \{ i \})$ (see for example Figure \ref{fig:importanceMatrices} on page \pageref{fig:importanceMatrices}) next to providing the values $D_l$ and their weighted average.

\paragraph{Special Case: Domain Shift Robustness}
If we understand one (or multiple) generative factor(s) $\bm{G}_S$ as indicating source domains which we would like to generalize over, we can use $\mathrm{PIDA}$ to evaluate robustness of a selected feature set $\bm{Z}_L$ against such domain shifts. In particular,
$$ \mathrm{IRS}(L | \{1,\dots,K\} \backslash \{S\}, \{S\}) $$
quantifies how robust $\bm{Z}_L$ is when changes in $\bm{G}_S$ occur. If we are building a model predicting a label $Y$ based on some (to be selected) feature set $L$, we can use this score to make a trade-off between robustness and predictive power. For example, we could use the best performing set of features among all those that satisfy a given robustness threshold.

\section{Estimation and Benchmarking Disentanglement}
\label{sec:samplingalgorithm}
In the supplementary material \ref{sec:estimation} we provide the derivation of our estimation procedure for $\mathrm{EMPIDA}(L| I, J)$. Here we only present the specific algorithm how $\mathrm{EMPIDA}$ can be estimated from a generic observational dataset $\mathcal{D}$ in Algorithm \ref{alg:empidaestimation}.  The main ingredient for this estimation to work is provided by our constrained causal model (i.e., a disentangled process) that implies that the backdoor criteria can be applied, which we showed in Proposition \ref{prop:properties}.

\begin{algorithm}[!htb]
\caption{EMPIDA Estimation}
\label{alg:empidaestimation}
\begin{algorithmic}[1]

\STATE \textbf{Input:} 
\STATE dataset $\mathcal{D} = \{ (\bm{x}^{(i)}, \bm{g}^{(i)}) \}_{i=1,\dots,N}$
\STATE trained encoder $E$
\STATE subsets of factors $L \subset \{1,\dots,K^\prime \}$ and $I, J \subset \{ 1,\dots,K \}$

\STATE \textbf{Preprocessing:}
\STATE encode all samples to obtain $\{\bm{z}^{(i)} = E(\bm{x}^{(i)}) : i = 1,\dots,N \}$ \label{lst:encoding}
\STATE estimate $p(\bm{g}^{(i)})$ and $p(\bm{g}^{(i)}_{\backslash (I \cup J)})$ $\forall i$ from relative frequencies in $\mathcal{D}$ \label{lst:frequencies}
%
\STATE \textbf{Estimation:}
\STATE find all realizations of $\bm{G}_I$ in $\mathcal{D}$: $\{ \bm{g}_I^{(k)}, k=1,\dots,N_I \}$\label{lst:partitionI1}
\STATE partition the dataset according to those realizations: \newline
$\mathcal{D}_I^{(k)} := \{ (\bm{x}, \bm{g}) \in \mathcal{D} \text{ s.t. } \bm{g}_I = \bm{g}_I^{(k)} \}$ \label{lst:partitionI2}
\FOR{$k = 1,\dots,N_I$}
\STATE estimate $\texttt{mean} \leftarrow \E [ \bm{Z}_L | \text{do}(\bm{G}_I \leftarrow \bm{g}_I^{(k)})]$ using Eq.~\eqref{eq:importancesampling} and samples $\mathcal{D}_I^{(k)}$ \label{lst:estimateMean}
\STATE partition $\mathcal{D}_I^{(k)}$ according to realizations of $\bm{G}_J$: \newline
$\mathcal{D}_{I,J}^{(k,l)} := \{ (\bm{x}, \bm{g}) \in \mathcal{D}_I^{(k)} \text{ s.t. } \bm{g}_J = \bm{g}_J^{(l)} \}$ \label{lst:partitionJ}
\STATE initialize $\texttt{mpida}(k) \leftarrow 0.0$
\FOR{$l = 1,\dots, N_{I,J}^{(k)}$}
\STATE $\texttt{mean}_\texttt{int} \leftarrow \E [ \bm{Z}_L | \text{do}(\bm{G}_I \leftarrow \bm{g}_I^{(k)}, \bm{G}_J \leftarrow \bm{g}_J^{(l)})]$ \newline
using Eq.~\eqref{eq:importancesampling} and samples $\mathcal{D}_{I,J}^{(k,l)}$ for estimation \label{lst:estimateMeanIntv}
\STATE compute $\texttt{pida} \leftarrow d(\texttt{mean}, \, \texttt{mean}_\texttt{int})$
\STATE update $\texttt{mpida}(k) \leftarrow \max \left( \texttt{mpida}(k),  \, \texttt{pida} \right)$
   \ENDFOR
   \ENDFOR
\STATE \textbf{Return} $\texttt{empida} \leftarrow \sum_{k=1}^{N_I} \frac{|\mathcal{D}_I^{(k)}|}{|\mathcal{D}|} \, \texttt{mpida}(k)$
\end{algorithmic}
\end{algorithm}

Even though the sampling procedure might look non-trivial at first sight, the algorithm \ref{alg:empidaestimation} for estimating  $\mathrm{EMPIDA}(L|I, J)$ has $\mathcal{O}(N)$ complexity as indicated by the following result:
\begin{proposition}[Computational Complexity]
\label{prop:complexity}
The $\mathrm{EMPIDA}$ estimation algorithm described in Algorithm \ref{alg:empidaestimation} scales $\mathcal{O}(N)$ in the dataset size $N = | \mathcal{D} |$.
\end{proposition}

The proof of Proposition \ref{prop:complexity} can be found in Appendix \ref{app:proof_complexit}. 
Note that a dataset capturing all possible variations generally grows exponentially in the number of generative factors. While this is a general issue for all validation approaches and care needs to be taken when collecting such datasets in practice, we just remind that due to the generally large nature of $N$ it is particularly important to have such an efficient validation procedure. In many benchmark datasets for disentanglement (e.g.\ dsprites) the observations are obtained noise-free and the dataset contains all possible combinations of generative factors exactly once. This makes the estimation of the disentanglement score even easier, as we have $| \mathcal{D}_{I=\{i\},J=\{ 1,\dots,K \} \backslash \{i\}}^{(k,l)} | = 1$. Furthermore, since no confounding is present, we can use conditioning to estimate the interventional effect, i.e., $p(\bm{x} | \text{do}(G_i \leftarrow g_i)) = p(\bm{x} | g_i)$, as seen in Proposition \ref{prop:properties} \ref{item:noconfounding}.
The disentanglement score of $Z_l$, as discussed in Eq.~\eqref{eq:disentanglementscore} , follows (see  \ref{sec:crossedataset} for details) as: 
$$ D_l = \max_{i \in \{1,\dots,K \}} \left( 1 - \frac{\mathrm{EMPIDA}_{li} }{\sup_{\tilde{\bm{x}} \in \mathcal{D}} d \left(\E [ Z_l], E( \tilde{\bm{x}} ) \right)} \right) .$$

\section{Experiments}\label{sec:experiments}

Our evaluations involve five different state of the art unsupervised disentanglement techniques (classic VAE, $\beta$-VAE, DIP-VAE, FactorVAE and $\beta$-TCVAE), each learning $10$ features.

\subsection{Methods Comparison}

In Table \ref{table:metrics} we provide a compact summary of our evaluation. Our objective is the analysis of various kinds of learned latent spaces and their characteristics, not primarily evaluating which methods work best under some metric. In particular, we used each method with the parameter settings that were indicated in the original publications (details are given in Appendix \ref{sec:experimentaldetails}) and did not tune them further in order to achieve a better robustness score, which is certainly feasible. Rather, we are interested in evaluating latent spaces as a whole, which encompasses both the method and its settings in combination. We can for example observe that $\beta$-TCVAE achieves a relatively low feature importance based measure by \citet{eastwood2018framework}. This is due to the fact that \citet{chen2018isolating} did not consider shape to be a generative factor in their tuning (which also leads to a lower informativeness score in our evaluation that includes this factor), and also because their model ends up with only few active dimensions. The treatment of such inactive components can make a difference when averaging disentanglement scores of the single $Z_i$ to an overall score. FI uses a simple average, MI weights the components with their overall feature importance and we weight them according to worst case deviation from mean (i.e., normalization of the $\mathrm{IRS}$).
\vspace{-5mm}
\setlength{\tabcolsep}{3pt}
\begin{table}[htb]
	\caption{\textbf{Metrics Overview:} IRS: (ours), FI: \citep{eastwood2018framework}, MI: \citep{ridgeway2018learning}, INFO: informativeness score \citep{eastwood2018framework} (higher is better). The number in parentheses indicates the rank according to a particular metric. Experimental details are given in Section \ref{sec:experimentaldetails}.}
	\label{table:metrics}
	\begin{center}
	\begin{tabular}{ l l l l l}
	  \textbf{Model} & \textbf{IRS} & \textbf{FI} & \textbf{MI} & \textbf{Info} \\
	  \hline \
	  VAE 									& $0.33$ $\color{gray}{(5)}$ & $0.23$ $\color{gray}{(4)}$ & $0.90$ $\color{gray}{(3)}$ & $0.82$ $\color{gray}{(1)}$\\
	  Annealed $\beta$-VAE 		& $0.57$ $\color{gray}{(2)}$ & $0.35$ $\color{gray}{(2)}$ & $0.86$ $\color{gray}{(5)}$ & $0.79$ $\color{gray}{(4)}$\\
	  DIP-VAE 							& $0.43$ $\color{gray}{(4)}$ & $0.39$ $\color{gray}{(1)}$ & $0.89$ $\color{gray}{(4)}$ & $0.82$ $\color{gray}{(1)}$\\
	  FactorVAE 							& $0.51$ $\color{gray}{(3)}$ & $0.31$ $\color{gray}{(3)}$ & $0.92$ $\color{gray}{(1)}$ & $0.79$ $\color{gray}{(4)}$ \\
	  $\beta$-TCVAE 				& $0.72$ $\color{gray}{(1)}$ & $0.16$ $\color{gray}{(5)}$ & $0.92$ $\color{gray}{(1)}$ & $0.74$ $\color{gray}{(5)}$ \\
	\end{tabular}
	\end{center}
\end{table}
\vspace{-5mm}

Believing that it is most insightful to look at scores for each dimension separately, which indicates the quality of a single feature, we included the full evaluations including plots of correspondance matrices (as in Figure \ref{fig:importanceMatrices}) in Appendix \ref{sec:vismatrices}. For future extensions and applications our work is added to the \texttt{disentanglement\textunderscore lib} of \citet{locatello2018challenging}. 

\begin{figure}[htb!]
\begin{center}
    \includegraphics[width=0.5\textwidth]{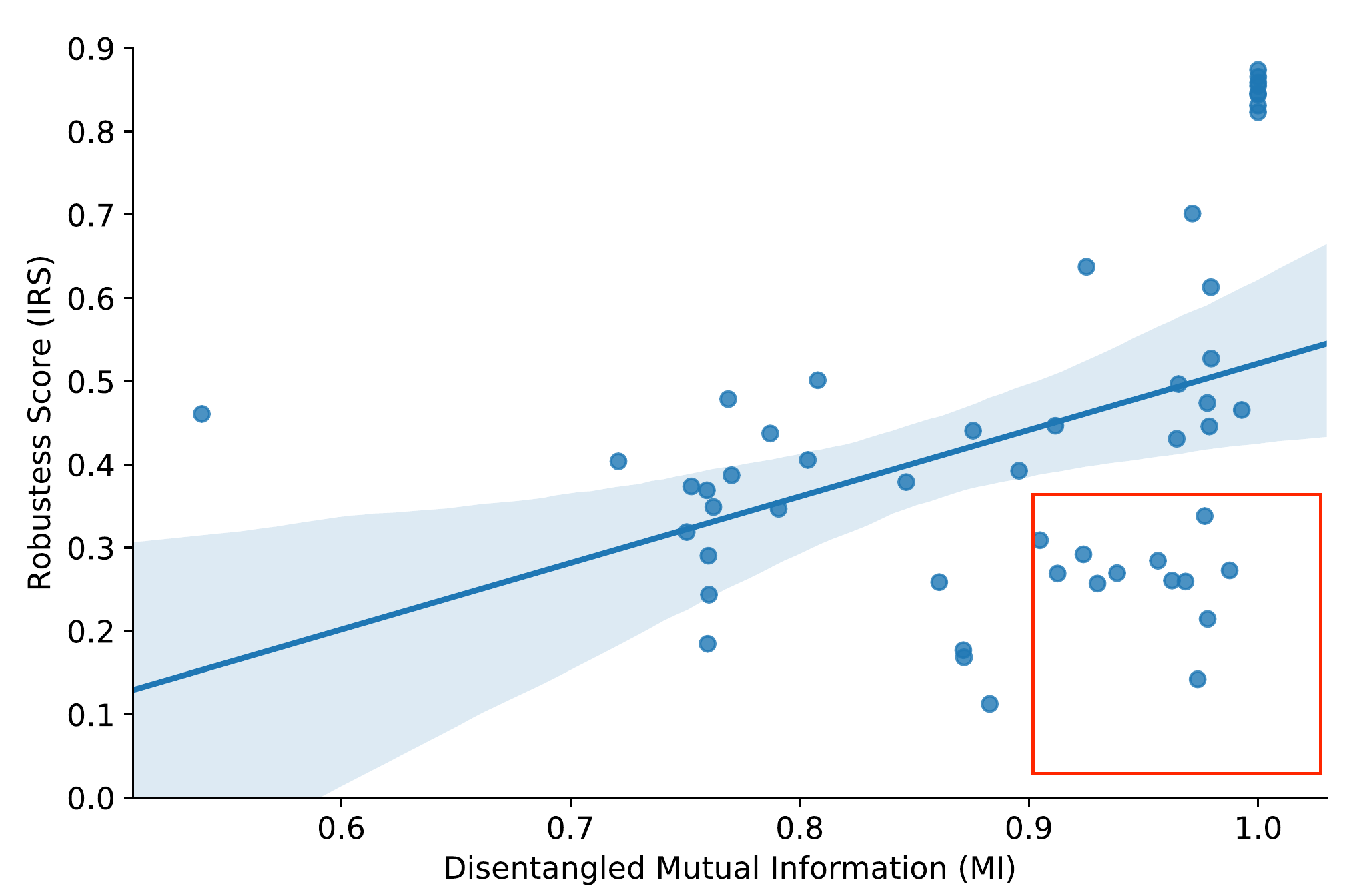}
  \caption{\textbf{Relationship Metrics:} Visualization of all learned features $Z_i$ in our universe (5 models with 10 dimensions each) based on their MI disentanglement score on the x axis and interventional robustness ($\mathrm{IRS}$) on the y axis. The red box indicates the features that obtained a high disentanglement score according to mutual information (i.e., they share high mutual information with only one generative factor), but still provide low robustness according to $\mathrm{IRS}$. These are the cases where the robustness perspective delivers additional insight into disentanglement quality.}
  \label{fig:relationshipMetrics}
 \end{center}
\end{figure}

\subsection{Robustness as Complementary Metric}
As we could already see in Table \ref{table:metrics}, different metrics do not always agree with each other about which model disentangles best. This is consistent with the recent large scale evaluation provided by \citet{locatello2018challenging}. In Figure \ref{fig:relationshipMetrics} we further illustrate the dependency between MI score and our IRS on the finer granularity of considering the metrics of individual features (instead of the full latent space). There seems to be a clear positive correlation between the two evaluation metrics. However, there are features classified as well disentangled according to MI, but not robustly (according to $\mathrm{IRS}$). These features are marked with the red rectangle in Figure \ref{fig:relationshipMetrics}.
We explore one typical such example in more detail in Figures \ref{fig:interventionalSpace}, \ref{fig:importanceMatrices} and \ref{fig:latentSpace} in the appendix, for the case of the DIP model. 
\begin{figure*}[htb!]
  \centering
    \includegraphics[width=1\textwidth, height=2.85cm]{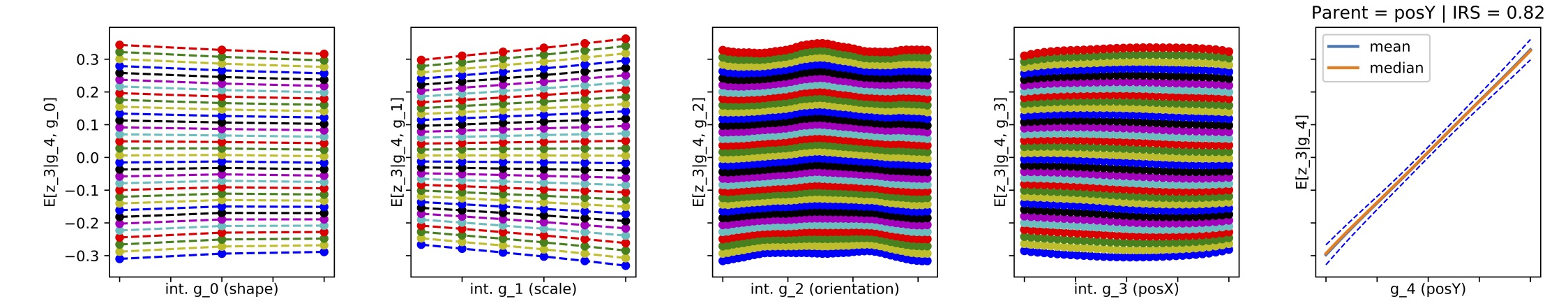}
    \includegraphics[width=1\textwidth, height=2.85cm]{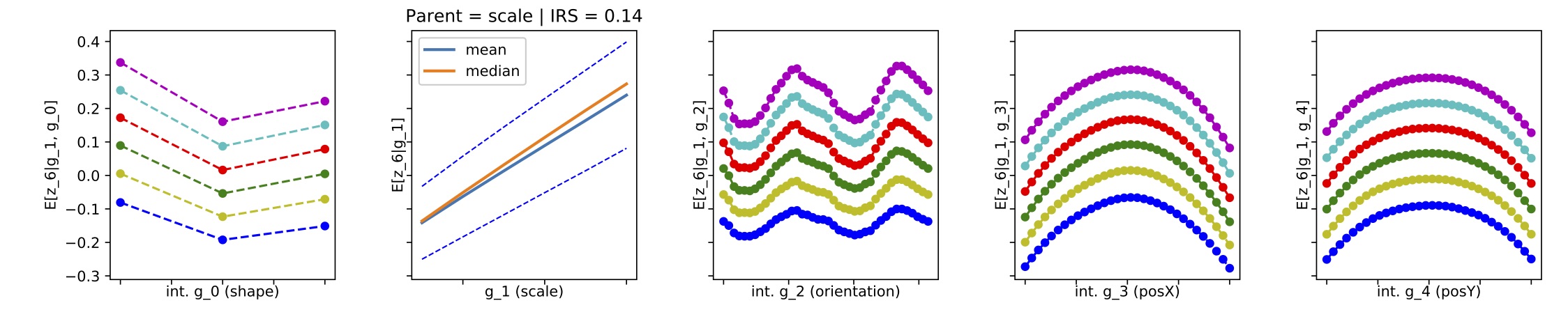}
  \caption{\textbf{Visualising Interventional Robustness:} Plots of $\E [Z_l | g_i*, \text{do}(G_j \leftarrow g_j^{\triangle})]$ as a function of $g_j^{\triangle}$ for different $G_j$ per column as explained in Section \ref{sec:visualizing}. The upper row is an example of good, robust disentanglement ($Z_3$ from the DIP model discussed in Figure \ref{fig:importanceMatrices}). The lower row illustrates $Z_6$ which is classified as well disentangled according to FI (top $18\%$) and MI (top $33\%$) but still has a low robustness score (bottom $4\%$). This stems from the fact that even though $Z_6$ is very informative about scale (almost a linear function in expectation), its value can still be changed remarkably by switching any of \texttt{posX}, \texttt{posY} or \texttt{orientation}. These additional dependencies are not discovered by mutual information (or feature importance) due to the higher noise in these relationships (see Figure \ref{fig:latentSpace}) and because they are partly hidden in cumulated effects. }
 \label{fig:interventionalSpace}
\end{figure*}

When there are rare events happening that still have a major impact on the features or when there is a cumulative effect from several generative factors (e.g., in Figure \ref{fig:interventionalSpace}), pairwise information based methods (such as MI or FI) cannot capture this vulnerability of deeply learned features. $\mathrm{IRS}$, on the other hand, looks specifically at these cases. For a well rounded view on disentanglement quality, we propose to use both types of measures in a manner that is complementary and use-case specific. Specificially when critical applications are designed on top of deep representations quantifying its robustness can be decisive.

\subsection{Visualising Interventional Robustness}\label{sec:visualizing}
We further introduce a new visualization technique for latent space models based on ground truth factors which is motivated by interventional robustness and illustrates how robust a learned feature is with respect to changes in nuisance factors. Figure \ref{fig:interventionalSpace} illustrates this approach on two features learned by the DIP model. Each row corresponds to a different feature $Z_l$. The upper row corresponds to a well disentangled and robust feature ($Z_3$) which gets classified as such by all three metrics. The lower ($Z_6$) also obtaines a high FI and MI score, however, $\mathrm{IRS}$ correctly discovers that this feature is not robust. This illustrates a case where having a robustness perspective on disentanglement is important.
The columns correspond to different generative factors $G_j$ (\texttt{shape}, \texttt{scale}, \texttt{orientation}, \texttt{posX}, \texttt{posY}) which potentially influence $Z_l$. For each latent variable $Z_l$ we first find the generative factor $G_{i^*}$ which is most related to it by choosing the maximizer of Eq.~\eqref{eq:disentanglementscore} (i.e., the factor that renders $Z_l$ most invariant). In the column $i^*$ we then plot the estimate of $\E [Z_l | g_{i^*}]$ together with its confidence bound in order to visualize the informativeness of $Z_l$ about $G_{i^*}$. For example the upper row in plot \ref{fig:interventionalSpace} corresponds to $Z_3$ in the DIP model and mostly relates to \texttt{posY}. This is why we plot the dependence of $Z_3$ on \texttt{posY} in the fifth column. The remaining columns then illustrate how $Z_3$ changes when interventions on the other generative factor are made, even though \texttt{posY} is being kept at a fixed value. Each line with different color corresponds to a particular value \texttt{posY} can take on. More generally speaking, we plot in the $j$th column $\E [Z_l | g_i*, \text{do}(G_j \leftarrow g_j^{\triangle})]$ as a function of $g_j^{\triangle}$ for all possible realizations $g_i*$ of $G_i*$. All values with constant $g_i*$ are connected with a line. For a robustly disentangled feature, we would expect all of these colored lines to be horizontal (i.e., there is no more dependency on any $G_j$ after accounting for $G_i*$).
As such visualizations can provide a much more in depth understanding of learned representations than single numbers, we provide the full plots of various models in the appendix \ref{sec:viseffects}.

\section{Conclusion}
We have proposed a framework for assessing disentanglement in deep representation learning which combines the generative process responsible for high dimensional observations with the subsequent feature encoding by a neural network. This perspective leads to a natural validation method, the \emph{interventional robustness score}. We show how it can be estimated from observational data using an efficient algorithm that scales linearly in the dataset size. As special cases, this proposed measure captures robust disentanglement and domain shift stability. 
Extensive evaluations showed that the existing metrics do not capture the effects that rare events or cumulative influences from multiple generative factors can have on feature encodings, while our robustness based validation metric discovers such vulnerabilities. 

We envision that the notion of interventional effects on encodings may give rise to the development of novel, robustly disentangled representation \emph{learning} algorithms, for example in the interactive learning environment \citep{thomas2017independently} or when weak forms of supervision are available \citep{bouchacourt2017multi, locatello2019disentangling}. The exploration of those ideas, especially including confounding, is left for future research.

\clearpage

\section*{Acknowledgments}
We thank Andreas Krause for helpful discussions and Alexander Neitz, Francesco Locatello and Olivier Bachem for the inclusion of our work into \texttt{disentanglement\textunderscore lib}  (\url{https://github.com/google-research/disentanglement_lib}) of \citet{locatello2018challenging}. This research was partially supported by the Max Planck ETH Center for Learning Systems.

\bibliography{bibliography}
\bibliographystyle{icml2019}

\clearpage

\appendix

\section{Estimation}\label{sec:estimation}
We now derive a sampling algorithm to estimate $\mathrm{IRS}$ from a observational dataset $\mathcal{D} = \{ \bm{x}^{(i)}, \bm{g}^{(i)} \}_{i=1,\dots,N}$ where $\bm{x}^{(i)} \in \mathbb{R}^n$ and  $\bm{g}^{(i)} \in \mathcal{G} = \mathcal{G}_1 \times \dots \times \mathcal{G}_K$ with each $\mathcal{G}_k$ being discrete and finite. In case of continous $\mathcal{G}_k$ we first need to perform a discretization. The discretization steps trade off bias and variance of the estimate through the number of samples that are available per combination of generative factors.

We will provide an estimation procedure for $\mathrm{EMPIDA}(L| I, J)$ as:
\begin{equation}\label{eq:empidacomplete}
\resizebox{.99\hsize}{!}{$
\E_{\bm{g}_I}  \left[ \sup_{\bm{g}_J^{\triangle}} \, d \left( \E [\bm{Z}_L | \text{do}(\bm{G}_I \leftarrow \bm{g}_I)], \, \E [\bm{Z}_L | \text{do}(\bm{G}_I \leftarrow \bm{g}_I, \bm{G}_J \leftarrow \bm{g}_J^{\triangle})]  \right) \right]$}.
\end{equation}
From that, also the $\mathrm{IRS}$ can be computed. In Section \ref{sec:crossedataset} we provide a simplified version that is sufficient for disentanglement benchmarking based on perfectly crossed noise free datasets. Readers most interested in this application might skip to that part.

The main ingredient for this estimation to work is provided by our constrained causal model (i.e., a disentangled process) that implies that the backdoor criteria can be applied, which we showed in Proposition \ref{prop:properties}.
Further, we already saw in Eq.~\eqref{eq:interventiononz} that $p(\bm{z}_L |   \bm{x}, \text{do}(\bm{G}_I \leftarrow \bm{g}_I, \bm{G}_J \leftarrow \bm{g}_J^{\triangle})) = p(\bm{z}_L |   \bm{x})$. This can be used to write the conditional expected value of $\bm{Z}_L$ as:

\resizebox{.99\linewidth}{!}{
  \begin{minipage}{\linewidth}
\begin{align}
& \E [\bm{Z}_L | \text{do}(\bm{G}_I \leftarrow \bm{g}_I, \bm{G}_J \leftarrow \bm{g}_J^{\triangle})] \nonumber \\
&= \quad \int \bm{z}_L \, p(\bm{z}_L |   \text{do}(\bm{G}_I \leftarrow \bm{g}_I, \bm{G}_J \leftarrow \bm{g}_J^{\triangle})) \, d \bm{z}_L \nonumber \\
&\overset{\mathclap{\eqref{eq:interventiononz}}}{=} \quad \int \int \bm{z}_L \, p(\bm{z}_L |   \bm{x}) \, p(\bm{x} |  \text{do}(\bm{G}_I \leftarrow \bm{g}_I, \bm{G}_J \leftarrow \bm{g}_J^{\triangle})) \, d \bm{x} \, d \bm{z}_L  \nonumber \\
&= \quad \int \left( \int \bm{z}_L \, p(\bm{z}_L |   \bm{x}) \, d \bm{z}_L \right) \, p(\bm{x} |  \text{do}(\bm{G}_I \leftarrow \bm{g}_I, \bm{G}_J \leftarrow \bm{g}_J^{\triangle})) \, d \bm{x} \nonumber \\
&\overset{\mathclap{\text{Prop.\ref{prop:properties}\ref{item:adjustmentset}}}}{=} \quad \int \left( \int \bm{z}_L \, p(\bm{z}_L |   \bm{x}) \, d \bm{z}_L \right) \, \nonumber \\ & \quad \quad  \left( \int p(\bm{x} |  \bm{g}_I, \bm{g}_J^{\triangle}, \bm{g}_{\backslash (I \cup J)}) \, p(\bm{g}_{\backslash (I \cup J)}) \, d\bm{g}_{\backslash (I \cup J)})\right) \, d \bm{x} \nonumber \\
&= \quad \int E(\bm{x})_L \, \left( \int p(\bm{x} |  \bm{g}_I, \bm{g}_J^{\triangle}, \bm{g}_{\backslash (I \cup J)}) \, p(\bm{g}_{\backslash (I \cup J)}) \, d\bm{g}_{\backslash (I \cup J)})\right) \, d \bm{x} \label{eq:estimateIntDist}
\end{align}
  \end{minipage}
}

where the elements $L$ of encoding $E(\cdot)$ are defined as:
$$ E(\bm{x})_L := \int \bm{z}_L \, q_\phi(\bm{z}_L |   \bm{x}) \, d \bm{z}_L. $$

It is now apparent how this formula can be used to estimate the expected value using the sample mean (or a robust alternative in case outliers in $\bm{x}$ are to be expected) based on a set of samples $\tilde{\mathcal{D}}$ drawn from $\int p(\bm{x} |  \bm{g}_I, \bm{g}_J^{\triangle}, \bm{g}_{\backslash (I \cup J)}) \, p(\bm{g}_{\backslash (I \cup J)}) \, d\bm{g}_{\backslash (I \cup J)}$ using the law of large numbers (LLN), i.e.,
\begin{equation}\label{eq:samplemean}
\E [ \bm{Z}_L | \text{do}(\bm{G}_I \leftarrow \bm{g}_I, \bm{G}_J \leftarrow \bm{g}_J^{\triangle})] \overset{\text{LLN}}{\approx} \frac{1}{| \tilde{\mathcal{D}} |} \sum_{\bm{x} \in \tilde{\mathcal{D}}} E(\bm{x})_L.
\end{equation}
However, all we are given are the samples $\mathcal{D}$ drawn from $p(\bm{x},\bm{g}) = p(\bm{x}|\bm{g}) \, p(\bm{g})$ where the generative factors could be confounded $p(\bm{g}) = \int p(\bm{g}|\bm{c}) p(\bm{c}) \, d\bm{c}$. 
This is why we now provide an importance sampling based adjusted estimation of the expected value of any function of the observations $h(\bm{X})$ after an intervention on $\bm{G}_J$ has occured and while conditioning on $\bm{G}_I$, i.e., $\E[ h(\bm{X}) | \text{do}(\bm{G}_I \leftarrow \bm{g}_I, \bm{G}_J \leftarrow \bm{g}_J^{\triangle})]$. This procedure can then be used to estimate Eq.~\eqref{eq:estimateIntDist}, as a special case with $h(\cdot) = E(\cdot)_L$, directly from $\mathcal{D}$.

By denoting the Kronecker-delta as $\delta$ we obtain:

\begin{align}\label{eq:importancesampling}
& \E_{\bm{X}}[ h(\bm{x}) | \text{do}(\bm{G}_I \leftarrow \bm{g}_I, \bm{G}_J \leftarrow \bm{g}_J^{\triangle})] \nonumber \\
&\overset{\mathclap{\text{derivation of } \eqref{eq:estimateIntDist}}}{=} \qquad \quad \int \int h(\bm{x}) \, p(\bm{x} |  \bm{g}_I, \bm{g}_J^{\triangle}, \bm{g}_{\backslash (I \cup J)}) \, \nonumber\\
 &\qquad
p(\bm{g}_{\backslash (I \cup J)}) \, d\bm{g}_{\backslash (I \cup J)} \, d \bm{x} \nonumber \\
&\overset{\mathclap{\bm{g}^{\prime} = (\bm{g}^{\prime}_I, \, \bm{g}^{\prime}_J, \, \bm{g}^{\prime}_{\backslash (I \cup J)})}}{=} \qquad \quad \int \int h(\bm{x}) \, p(\bm{x} |  \bm{g}^\prime) \, p(\bm{g}_{\backslash (I \cup J)}^\prime) \, \delta(\bm{g}_I^\prime - \bm{g}_I) \, \nonumber\\
& \qquad\delta (\bm{g}_J^\prime - \bm{g}_J^\triangle) \, d\bm{g}^\prime \, d \bm{x} \nonumber \\
&\overset{\mathclap{\text{LLN}}}{\approx} \qquad \quad \frac{1}{N} \sum_{\bm{x}^{(i)}, \, \bm{g}^{(i)} \in \mathcal{D}} h(\bm{x}^{(i)})  \nonumber\\
&\quad\quad  \frac{p(\bm{x}^{(i)}| \bm{g}^{(i)}) p(\bm{g}_{\backslash (I \cup J)}^{(i)}) \delta(\bm{g}_I^{(i)} - \bm{g}_I) \, \delta (\bm{g}_J^{(i)} - \bm{g}_J^\triangle)}{p(\bm{x}^{(i)}| \bm{g}^{(i)}) p(\bm{g}^{(i)})} \nonumber \\
&= \qquad \quad \sum_{\substack{\bm{x}^{(i)}, \, \bm{g}^{(i)} \in \mathcal{D} \\ \text{where } \bm{g}_I^{(i)}=\bm{g}_I \\ \text{and } \bm{g}_J^{(i)}=\bm{g}_J}} h(\bm{x}^{(i)}) \frac{p(\bm{g}_{\backslash (I \cup J)}^{(i)}) }{N \, p(\bm{g}^{(i)})}
\end{align}
We can rewrite the weighting term as:
$$w_i := \frac{p(\bm{g}_{\backslash (I \cup J)}^{(i)}) }{N \, p(\bm{g}^{(i)})} = \frac{1}{N \, p(\bm{g}_I^{(i)}, \bm{g}_J^{(i)}  |\bm{g}_{\backslash (I \cup J)}^{(i)})}$$
which gives us the natural interpretation that samples $\bm{g}_I, \bm{g}_J$ that would occur more often together with a certain $\bm{g}_{\backslash (I \cup J)}$ need to be downweighted in order to correct for the confounding effects. We can also see that in case of statistical independence between the generative factors, this reweighting is not needed and we can simply use the sample mean with the subselection of the dataset $\mathcal{D}_{sel} = \{ (\bm{x}^{(i)}, \, \bm{g}^{(i)}) \in \mathcal{D} : \bm{g}_I^{(i)}=\bm{g}_I \text{ and } \bm{g}_J^{(i)}=\bm{g}_J^{\triangle} \}$.

Since we assume $\bm{G}$ to be discrete, we can estimate these reweighting factors $w_i$ from observed frequencies. Even though this sampling procedure looks non-trivial, we show in Section \ref{sec:samplingalgorithm} how it can be used to obtain an $\mathcal{O}(N)$ estimation algorithm for $\mathrm{EMPIDA}(L|I, J)$.

\subsection{Crossed Dataset without Noise: Benchmarking Disentanglement}\label{sec:crossedataset}
In many benchmark datasets for disentanglement (e.g.\ dsprites) the observations are obtained noise free and the dataset contains all possible crossings of generative factors exactly ones. This makes the estimation of the disentanglement score very efficient, as we have $| \mathcal{D}_{I=\{i\},J=\{ 1,\dots,K \} \backslash \{i\}}^{(k,l)} | = 1$. Furthermore, since no confounding is present, we can use conditioning to estimate the interventional effect, i.e., $p(\bm{x} | \text{do}(G_i \leftarrow g_i)) = p(\bm{x} | g_i)$, as seen in Proposition \ref{prop:properties} \ref{item:noconfounding}.
In order to obtain the disentanglement score of $Z_l$, as discussed in Eq.~\eqref{eq:disentanglementscore}, we therefore just need to compute the $\mathrm{PIDA}$ value:
$$ d ( \E [ \bm{Z}_l | g_i^{(k)}], E(\tilde{\bm{x}})_l ) \quad \forall \tilde{\bm{x}} \in \mathcal{D}_i^{(k)} $$
for all generative factors $G_i$ and realizations thereof $\{g_i^{(1)},\dots, g_i^{(N_i)} \}$. $\mathcal{D}_i^{(k)}$ is the set of observations that was generated with a particular configuration $g_i^{(k)}$.
We choose the maximum value w.r.t.\ $\tilde{\bm{x}}$ as $\mathrm{MPIDA}$ and average over realizations $g_i^{(k)}$ to obtain:
\begin{align*}
 \mathrm{EMPIDA}_{li} :&= \mathrm{EMPIDA}(\{l \} | \{i\}, \{ 1,\dots,K \} \backslash \{i\})  \\&=  \frac{1}{N_i} \sum_{k=1}^{N_i} \sup_{\tilde{\bm{x}} \in \mathcal{D}_i^{(k)}} d \left(\E [ Z_l | g_i^{(k)}], E( \tilde{\bm{x}} ) \right).
\end{align*}

The estimate for the disentanglement score in Eq.~\eqref{eq:disentanglementscore} for $Z_l$ follows from that:
$$ D_l = \max_{i \in \{1,\dots,K \}} \left( 1 - \frac{\mathrm{EMPIDA}_{li} }{\sup_{\tilde{\bm{x}} \in \mathcal{D}} d \left(\E [ Z_l], E( \tilde{\bm{x}} ) \right)} \right) .$$

\section{Proof of Proposition \ref{prop:properties} }
\label{app:proof}

\begin{proof}
\label{proof}
Property \ref{item:causalmechanism} directly follows from Definition \ref{def:disentangledmodel} and the definition of an independent causal mechanism. \ref{item:dependentz} and \ref{item:dependentzgivenx} can be read off the graphical model \citep{koller2009probabilistic} in Figure \ref{fig:graphicalmodel} which does not contain any arrow from $G_i$ to $G_j$ for $i\neq j$ by Definition \ref{def:disentangledmodel} of the constrained SCM. This is due to the fact that any distribution implied by an SCM is Markovian with respect to the corresponding graph \citep[Prop. 6.31]{peters2017elements}.
\ref{item:processinginequality} follows from the data processing inequality since we have $\bm{X} \independent C | \bm{G}$.
The non-existence of a directed path from $G_j$ to $G_i$ implies that there is no total causal effect \citep[Prop. 6.14]{peters2017elements}. This, in turn, is equivalent to property \ref{item:interventiononz} \citep[Prop. 6.13]{peters2017elements}. 
Finally, since there are no arrows between the $G_i$'s, the backdoor criterion \citep[Prop. 6.41]{peters2017elements} can be applied to estimate the interventional effects in \ref{item:adjustmentset}. In particular, $\bm{G}_{\backslash j}$ blocks all paths from $G_j$ to $\bm{X}$ entering $G_j$ through the backdoor (i.e., $G_j \leftarrow \dots \rightarrow \bm{X}$) but at the same time does not contain any descendents of $G_j$ since by definition $G_j \not\rightarrow G_i$ $\forall i \neq j$. Property \ref{item:noconfounding} also follows from $G_j \not\rightarrow G_i$ $\forall i \neq j$ by using parent adjustment \citep[Prop. 6.41]{peters2017elements}, where in the case no confounding $\bm{PA}_j = \emptyset$.
These properties is why the constrained SCM in Definition \ref{def:disentangledmodel}  is important for further estimation.
\end{proof}

\section{Proof of Proposition \ref{prop:complexity} }
\label{app:proof_complexit}

\begin{proof}
The encodings in line \ref{lst:encoding} requires one pass through the dataset $\mathcal{D}$. So does the estimation of the occurance frequencies in line \ref{lst:frequencies} as one can use a hash table to keep track of the number of occurances of each possible realization. Therefore, the preprocessing steps scale with $\mathcal{O}(N)$.

Further, also the partitioning of the full dataset into $\mathcal{D} = \bigcup_{k=1}^{N_I} \bigcup_{l=1}^{N_{I,J}^{(k)}} \mathcal{D}_{I,J}^{(k,l)}$, which is done in lines \ref{lst:partitionI1}, \ref{lst:partitionI2} and \ref{lst:partitionJ}, can be done with two passes through the dataset by using hash tables: In the first pass we create buckets with $\bm{g}_I^{(k)}$ as keys. Consequently, we can pass through all of these buckets to create subbuckets where $\bm{g}_J^{(l)}$ is used as key. This reasoning is further illustrated in Figure \ref{fig:partitioning} and leads us to the $\mathcal{O}(N)$ complexity of the partitioning.

The remaining computational bottleneck are the computations of $\texttt{mean}$ in line \ref{lst:estimateMean} and $\texttt{mean}_{\texttt{intv}}$ in line \ref{lst:estimateMeanIntv}.
Using Eq.~\eqref{eq:importancesampling} we obtain $\E [ \bm{Z}_L | \text{do}(\bm{G}_I \leftarrow \bm{g}_I^{(k)})] \approx \sum_{\bm{x}^{(i)} \in \mathcal{D}_I^{(k)}} w_i \, E(\bm{x}^{(i)})$ to compute \texttt{mean} and $\E [ \bm{Z}_L | \text{do}( \bm{G}_I \leftarrow \bm{g}_I^{(k)}, \bm{G}_J \leftarrow \bm{g}_J^{(l)})] \approx \sum_{\bm{x}^{(i)} \in \mathcal{D}_{I,J}^{(k,l)}} \tilde{w}_i \, E(\bm{x}^{(i)})$ to compute $\texttt{mean}_{\texttt{intv}}$. Since we already computed the encodings as well as the reweighting terms in the preprocessing step, these summations scale as $\mathcal{O}(|\mathcal{D}_I^{(k)}|)$ and $\mathcal{O}(|\mathcal{D}_{I,J}^{(k,l)}|)$. As can be seen in Figure \ref{fig:partitioning}, it holds that $\sum_{k=1}^{N_I} |\mathcal{D}_I^{(k)}| = N$ as well as $\sum_{k=1}^{N_I} \sum_{l=1}^{N_{I,J}^{(k)}} | \mathcal{D}_{I,J}^{(k,l)} | = N$ which implies the total computational complexity of $\mathcal{O}(N)$.
\end{proof}

\begin{figure*}
\begin{tabular}{| p{3cm} | p{3cm} | p{3cm} | p{1cm} p{1cm} p{1cm}}
  \cline{1-3}
  $\bm{G}_I$ fixed & $\bm{G}_J$ fixed & remaining $\bm{G}_{\backslash (I \cup J)}$ \\
  \cline{1-3} \cline{1-3}
    								& $\bm{g}_J^{(1)}$ & $\mathcal{D}_{I,J}^{(1,1)}$ & & & \rdelim\}{20}{5mm}[$|\mathcal{D}| = N$]\\[0.5cm] \cline{2-3}
   	$\bm{g}_I^{(1)}$	& $\cdots$ & $\cdots$ \\[0.5cm] \cline{2-3}
   									& $\bm{g}_J^{(N_{I,J}^{(1)})}$ & $\mathcal{D}_{I,J}^{(1,N_{I,J}^{(1)})}$ \\[0.5cm]
  \cline{1-3}
      								& $\cdots$ & $\cdots$ & \rdelim\}{2}{6mm}[$| \mathcal{D}_{I,J}^{(k,l)} |$] & \rdelim\}{6}{6mm}[$| \mathcal{D}_{I}^{(k)} |$]  \\[0.5cm] \cline{2-3}
   	$\cdots$	& $\cdots$ & $\cdots$ \\[0.5cm] \cline{2-3}
   									& $\cdots$ & $\cdots$ \\[0.5cm]
   \cline{1-3}
       								& $\bm{g}_J^{(1)}$ & $\mathcal{D}_{I,J}^{(N_I,1)}$ \\[0.5cm] \cline{2-3} 
   	$\bm{g}_I^{(N_I)}$	& $\cdots$ & $\cdots$ \\[0.5cm] \cline{2-3}
   									& $\bm{g}_J^{(N_{I,J}^{(N_I)})}$ & $\mathcal{D}_{I,J}^{(N_I,N_{I,J}^{(N_I)})}$ \\[0.5cm]
   	\cline{1-3}
\end{tabular}
\caption{\textbf{Partitioning of Dataset:} In order to estimate $\mathrm{EMPIDA}(L | I, J)$ we first partition the dataset according to possible realizations of $\bm{G}_I$ (first column), where we assume there are $N_I$ many. This partitioning can be done in linear time $\mathcal{O}(N)$ by using hash tables with $\bm{g}_I^{(i)}$ as keys. For each such partition $D_I^{(k)}$ we can further split these sub-datasets according to realizations of $\bm{G}_J$ to obtain $\mathcal{D}_{I,J}^{(k,l)} = \{ (\bm{x}, \bm{g}) \in \mathcal{D} \text{ s.t. } \bm{g}_I = \bm{g}_I^{(k)}, \bm{g}_J = \bm{g}_J^{(l)} \}, \, l=1,\dots,N_I^{(k)}$ (illustrated as boxes in third column). We denote with $N_I^{(k)}$ the number of realizations of $\bm{G}_J$ that occur together with $\bm{g}_I^{(k)}$ (i.e., can be found in $\mathcal{D}_I^{(k)}$). This takes $\mathcal{O}(| D_I^{(k)} |)$ time per partition $D_I^{(k)}$ or $\mathcal{O}(\sum_{k=1}^{N_I} | D_I^{(k)} |) = \mathcal{O}(N)$ in total by again making use of hash tables.}
\label{fig:partitioning}
\end{figure*}

\paragraph{Real World Considerations:} Though this estimation procedure scales $\mathcal{O}(N)$ in the dataset size, the required number of observations for a fixed estimation quality (i.e., if $|\mathcal{D}_{I,J}^{(k,l)}|$ should stay constant) might become very large, as we have exponentially growing (in $|I|$ and $|J|$) many possible combinations to consider. This is why some trade-offs need to be made when comparing large sets of factors. The estimation for $|I|, |J| = 1, 2 \text{ or } 3$, however, usually works well. One trade-off parameter is the discretization step of of $g_i$'s. Partitioning a factor into fewer realizations yields less possible combinations and hence larger sets $\mathcal{D}_{I,J}^{(k,l)}$. In general, the more noise we expect in $\bm{x}$ the larger the sets $\mathcal{D}_{I,J}^{(k,l)}$ we want to have in order to obtain stable estimates of the expected values. Also, if we allow for fewer possible realizations in the generative factors, the smaller our dataset can be to cover all relevant combinations. However, larger discretization steps come at the cost of having a less sensitive score.
Also note that taking the supremum is in general not vulnerable to outliers in $\bm{x}$ as we compute distances of \emph{expected} values. When outliers are to be expected, a robust estimate for these expected values can be used. Only when little data is available special care needs to be taken.

\section{Details of Experimental Setup}\label{sec:experimentaldetails}

\subsection{Validation Methods}
We compute the feature importance based disentanglement scores, as discussed by \citet{eastwood2018framework}, using random forests with 50 decision trees that are split up to a minimal leaf size of 500. As opposed to \citet{eastwood2018framework}, we only use one single feature to 'randomly choose from' at each split, since this guarantees that each feature is equally given the chance to prove itself in reducing the out-of-bag error. When multiple features can be chosen from at each split, it is well possible that features with a mediocre importance are never chosen as there are features always yielding a better split. This would lead to an underestimation of their importance.

For the mutual information metric \citep{ridgeway2018learning} we followed the original proposal of discretizing each latent dimension into 20 buckets and computing the discrete mutual information based on that. We found that using smaller discretization steps (i.e., more buckets) does not change the results notably.

Since we make comparisons to information based evaluation methodologies by \citet{eastwood2018framework} and \citet{ridgeway2018learning}, we here give a more in depth overview of these methods. The validation method of \citet{eastwood2018framework} is based on training a predictor model (e.g.\ a random forest) which tries to estimate the true generative factors based on the latent encoding. The way disentanglement can be observed is by analyzing the feature importances implicit in this regressor. Intuitively, we expect that in a disentangled representation, each dimension contains information about one single generative factor.
In particular, \citet{eastwood2018framework} proceed as follows:
Given a labeled dataset with generative factors and observations $\mathcal{D} = \{ \bm{x}^{(i)}, \bm{g}^{(i)} \}_{i=1,\dots,N}$ and a given encoder $E$ (to be evaluated), they first create the set of features $\{\bm{z}^{(i)} = E(\bm{x}^{(i)}) \in \mathbb{R}^{K^\prime} : i = 1,\dots,N \}$.
Using these $K^\prime$ features as predictors, they train an individual regressor $f_i$ for each generative factor $G_i$, i.e., $\hat{G}_i = f_i(\bm{Z})$. 
As the basis for further computations, they set up a matrix of relative importances $\bm{R}$ based on these feature importance values. In particular, $R_{ij}$ denotes the relative importance of the feature $Z_i$ when predicting $G_j$.

\begin{figure*}[htb!]
  \centering
    \includegraphics[width=1.0\textwidth]{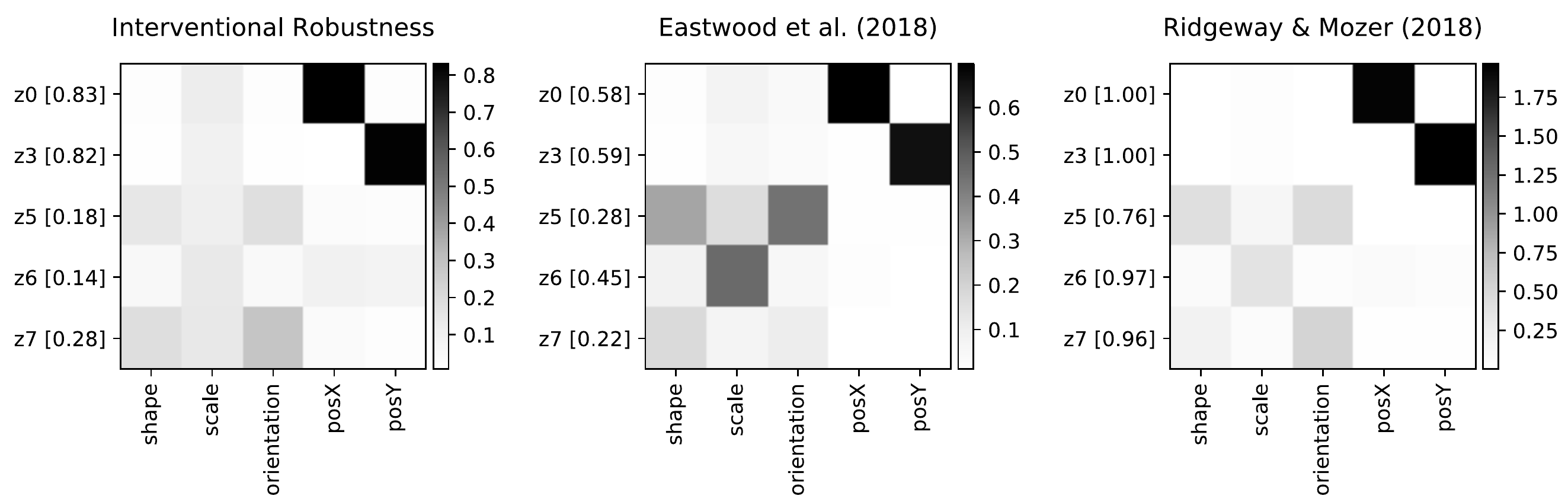}
  \caption{\textbf{Dependency Matrices:} These plots illustrate the different dependency structure matrices (of the features learned by the DIP model) that are used by the three discussed evaluation metrics. The rows correspond to the latent space dimensions $Z_i$ (disentanglement score of each feature is given in brackets) and the columns to generative factors $G_j$ (labels indicates their interpretation in the dsprites dataset).}
  \label{fig:importanceMatrices}
\end{figure*}

Plotting the matrix $\bm{R}$ gives a good first impression of the disentanglement capabilities of an encoder. Ideally, we would want to see only one large value per row while the remaining entries should be zero.
In our experimental evaluations we plot this matrix (together with similarly interpretable matrices of the other metrics) as is shown for example in Figure \ref{fig:importanceMatrices} on page \pageref{fig:importanceMatrices}.

To explicitly quantify this visual perspective, \citet{eastwood2018framework} summarize disentanglement as one score value which measures to what extent indeed each latent dimension can only be used to predict one generative factor (i.e., sparse rows). It is obtained by first computing the `probabilities' of $Z_i$ being important to predict $G_j$,
\begin{equation*}
P_{ij} = R_{ij} / \sum_{k=0}^{K-1} R_{ik}
\end{equation*}
and the entropy of this distribution: $H_K(P_{i\cdot}) = - \sum_{k=0}^{K-1} P_{ik} \log_K P_{ik}$, where $K=\text{dim} (\bm{g})$ is the number of generative factors. The disentanglement score of variable $Z_i$ is then defined as 
$D_i = (1 - H_K(P_{i\cdot})).$ For example, if only one generative factor $G_u$ can be predicted with $Z_i$, i.e., $P_{ij} = \delta_{iu}$, we obtain $D_i = 1$. If the explanatory power spreads over all factors equally, the score is zero. Using relative variable importance $\rho_i = \sum_j R_{ij} / \sum_{ij} R_{ij}$, which accounts for dead or irrelevant components in $\bm{Z}$, they find an overall disentanglement score as weighted average $S_D = \sum_i \rho_i D_i$. When later plotting the full importance matrices, we also provide information about the individual feature disentanglement scores $D_i$ in the corresponding row labels. These feature-wise scores are better comparable between metrics since all of them have different heuristics to obtain the (weighted) average $S_D$.

As an additional measure to obtain a more complete picture of the quality of the learned code, they additionally propose the \textit{informativeness} score. It tells us how much information about the generative factors is captured in the latent space and is computed as the out-of-bag prediction accuracy of the regressors $f_1,\dots,f_K$. In our evaluations in Section \ref{sec:experiments} we will also provide this score, as there is often a trade-off between a disentangled structure and information being preserved.

The mutual information based metric by \citet{ridgeway2018learning} proceeds in a similar way to \citet{eastwood2018framework}. However, instead of relying on a random forest to compute the feature importances, they use an estimate of the mutual information between encodings and generative factors. In particular, they also first compute an importance matrix $\tilde{\bm{R}}$ where the element $\tilde{R}_{ij}$ corresponds to the mutual information between $Z_i$ and $G_j$. We also provide plots of this matrix whenever evaluations are made (e.g.\ Figure \ref{fig:importanceMatrices} on page \pageref{fig:importanceMatrices}). Another difference to \citet{eastwood2018framework} is that \citet{ridgeway2018learning} do not compute entropies to measure the deviation from the ideal case of having only one large value per row. Instead, they compute a normalized squared difference between each row and its idealized case where all values except the largest are set to zero. To summarize the disentanglement scores of different dimensions in a feature space they use an unweighted average. 

\subsection{Disentanglement Approaches}
For the disentangling VAE models we made use of existing implementations where this was available. Classic VAE \citep{kingma2013auto} and DIP-VAE \citep{kumar2017variational} we implemented ourselves and trained them for $300$ epochs using Adam \citep{kingma2014adam} with a learning rate of \texttt{1e-4} and batch size of $512$. We used the same neural network architecture as is described in the appendix of \citet{chen2018isolating}. For DIP-VAE we set the parameters to $\lambda_d=100, \lambda{od}=10$, as is used in the original publication.
For the annealed $\beta$-VAE approach \citep{burgess2018understanding} we used the publicly available third party code from \texttt{https://github.com/1Konny/Beta-VAE}, where parameters are set to $C=20$ and $\gamma=100$. Also, for FactorVAE \citep{kim2018disentangling} we used third party code from \texttt{https://github.com/1Konny/FactorVAE} with their parameter $\gamma=6.4$.
\citet{chen2018isolating} provided their own code for $\beta$-TCVAE at \texttt{https://github.com/rtqichen/beta-tcvae}, which we made use of. We kept their chosen default parameters ($\beta=6.0$).

\section{Visualisations of Importance Matrices}\label{sec:vismatrices}
\begin{figure*}[htb!]
  \centering
    \includegraphics[width=1\textwidth]{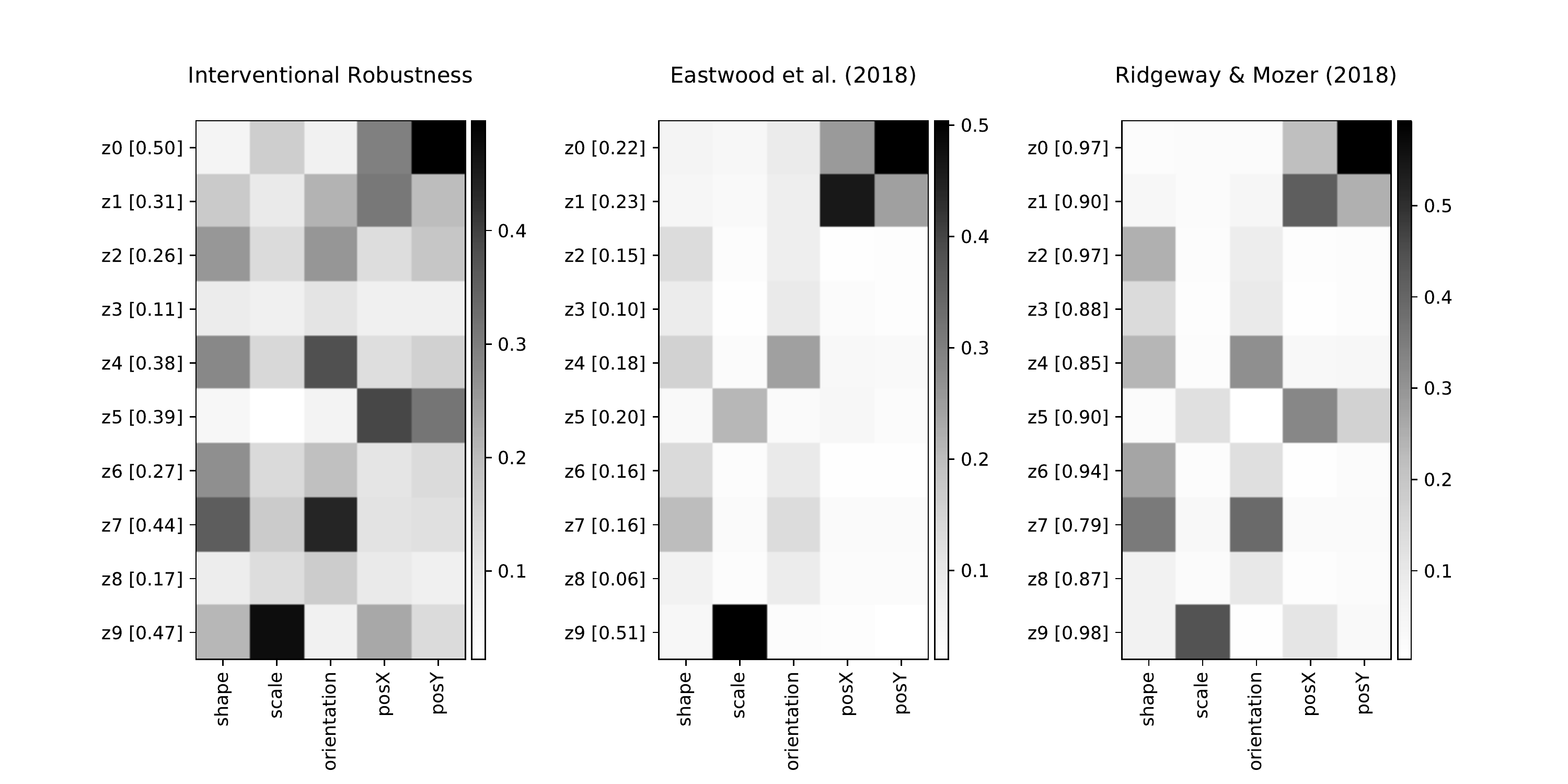}
  \caption{Importance matrices of all three validation metrics for the classic VAE model \citep{kingma2013auto}.}
  \label{fig:matrixVAE}
\end{figure*}

\begin{figure*}[htb!]
  \centering
    \includegraphics[width=1\textwidth]{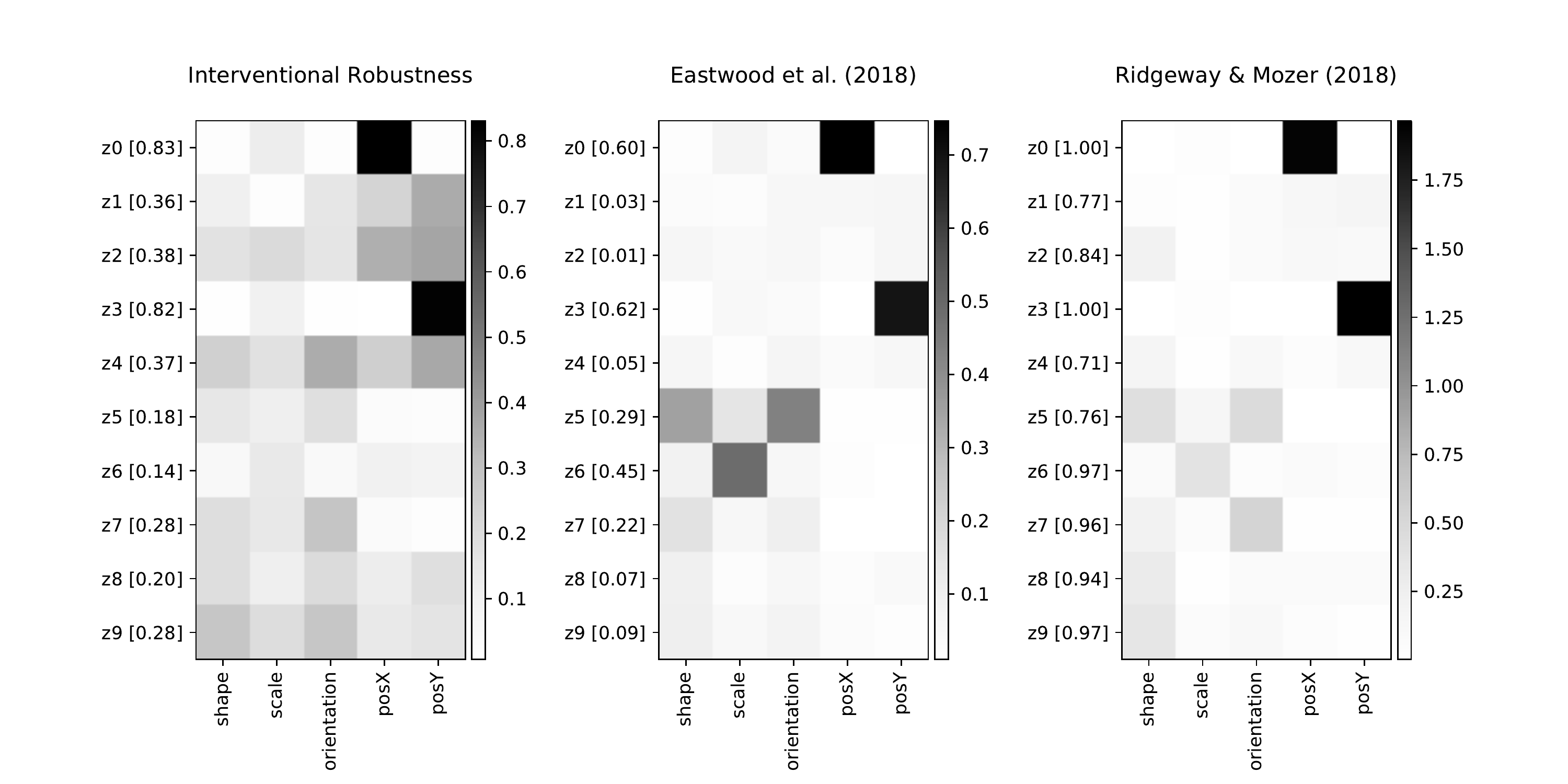}
  \caption{Importance matrices of all three validation metrics for the DIP-VAE model \citep{kumar2017variational}.}
  \label{fig:matrixDIP}
\end{figure*}

\begin{figure*}[htb!]
  \centering
    \includegraphics[width=1\textwidth]{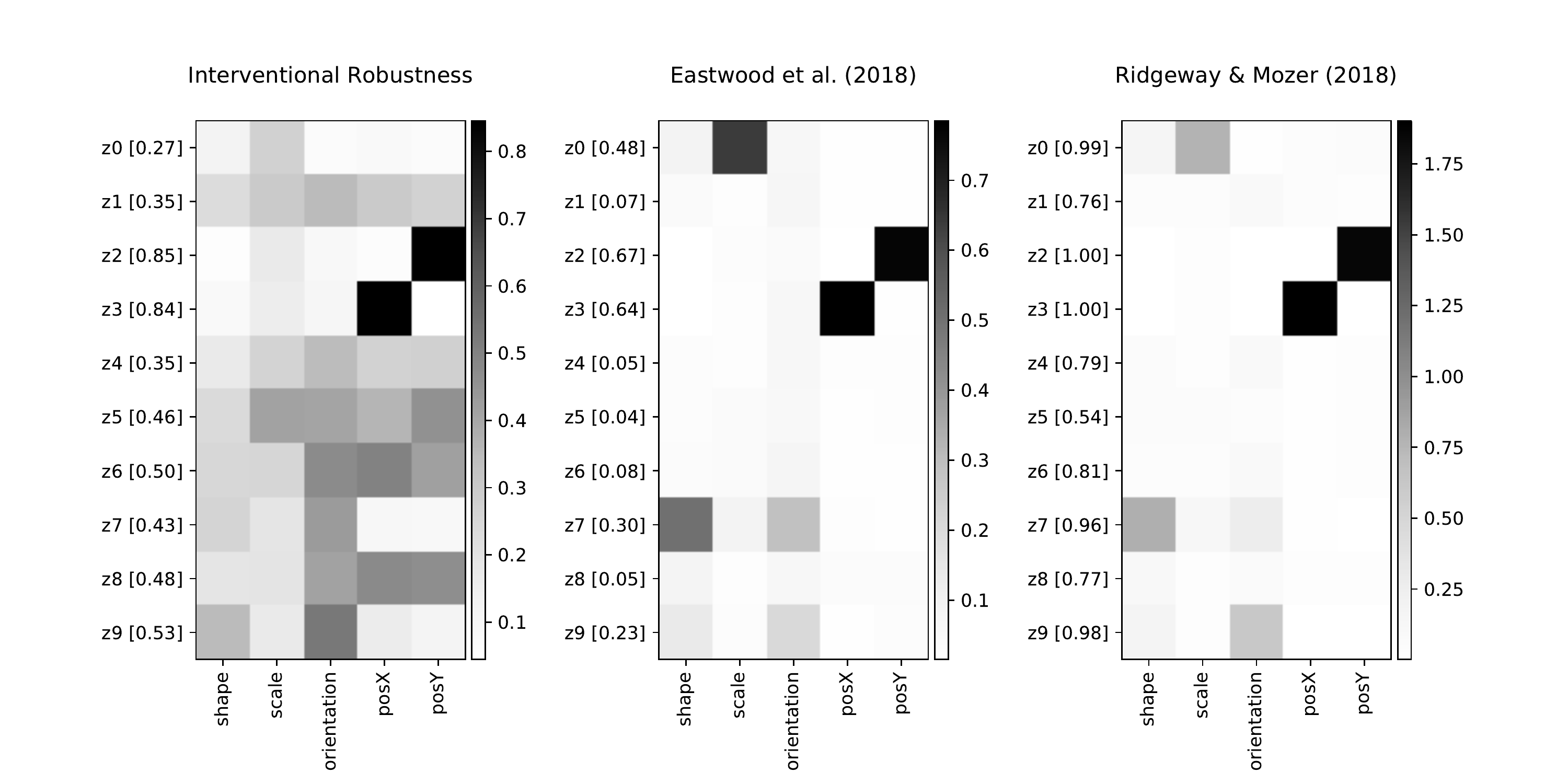}
  \caption{Importance matrices of all three validation metrics for the annealed $\beta$-VAE model \citep{higgins2016beta, burgess2018understanding}.}
  \label{fig:matrixannealedBeta}
\end{figure*}

\begin{figure*}[htb!]
  \centering
    \includegraphics[width=1\textwidth]{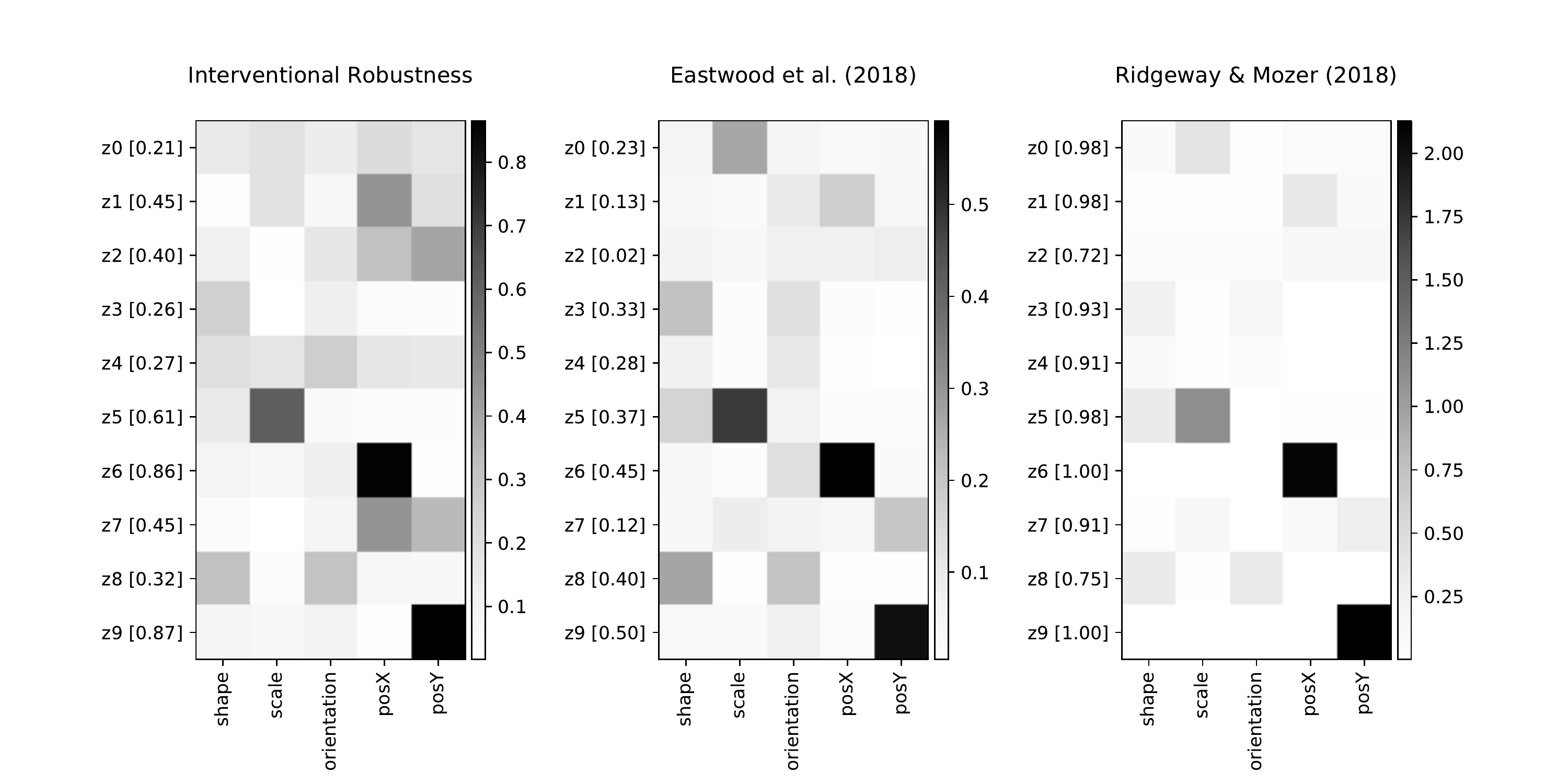}
  \caption{Importance matrices of all three validation metrics for the FactorVAE model \citep{kim2018disentangling}.}
  \label{fig:matrixFactorVAE}
\end{figure*}

\begin{figure*}[htb!]
  \centering
    \includegraphics[width=1\textwidth]{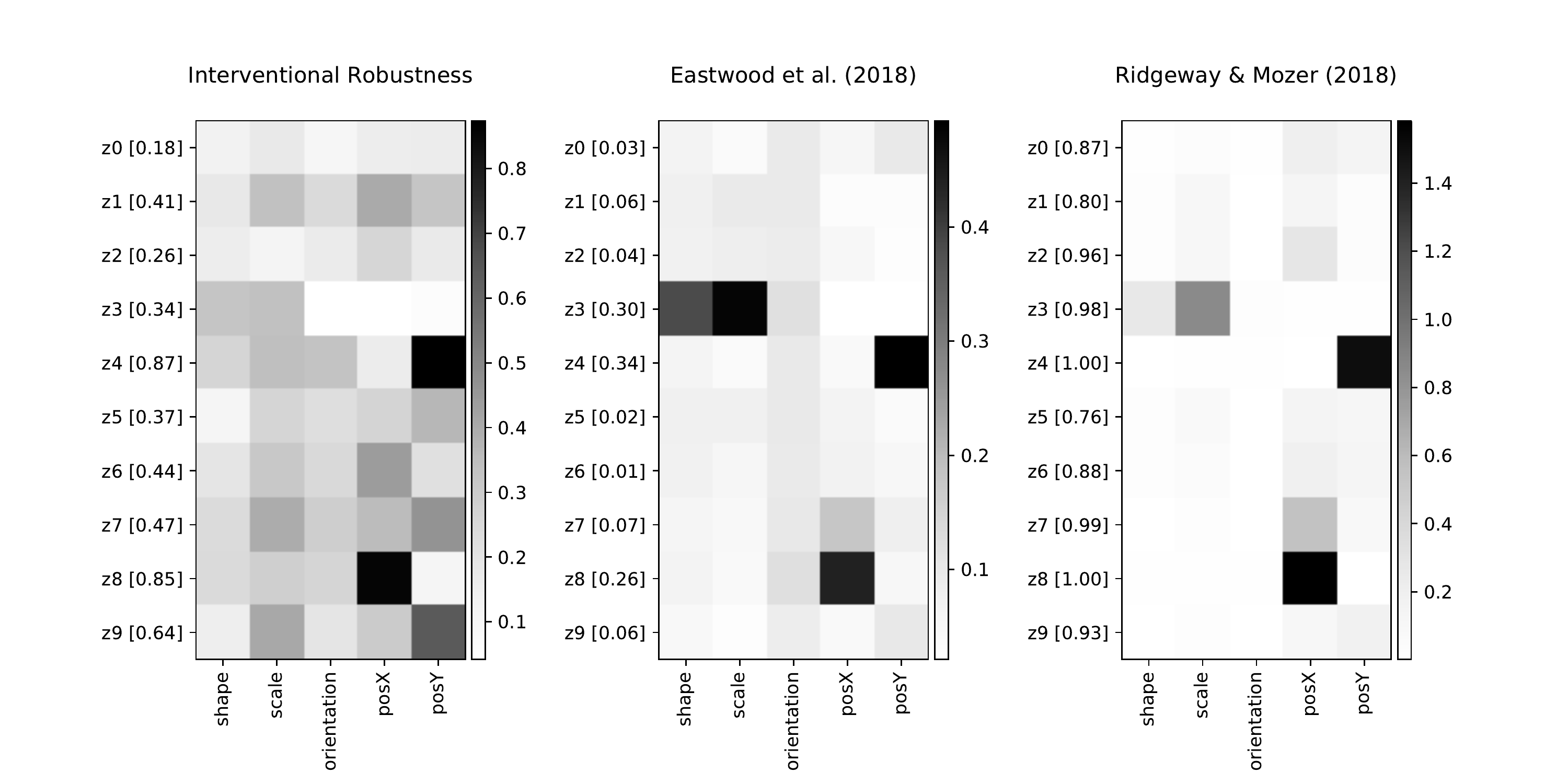}
  \caption{Importance matrices of all three validation metrics for the $\beta$-TCVAE model \citep{chen2018isolating}.}
  \label{fig:matrixbetatcvae}
\end{figure*}

Plots of the full importance matrices for the considered latent spaces and all three validation metrics are included in Figures \ref{fig:matrixVAE}, \ref{fig:matrixDIP}, \ref{fig:matrixannealedBeta}, \ref{fig:matrixFactorVAE} and \ref{fig:matrixbetatcvae}. The y labels include the disentanglement scores of each individual feature $Z_i$.

\begin{figure*}[htb!]
  \centering
    \includegraphics[width=0.95\textwidth]{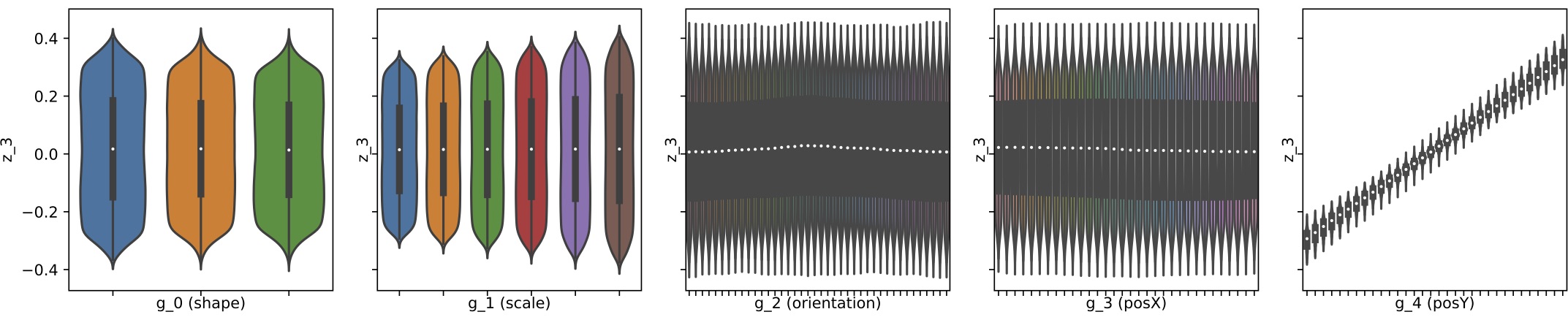}
    \includegraphics[width=0.95\textwidth]{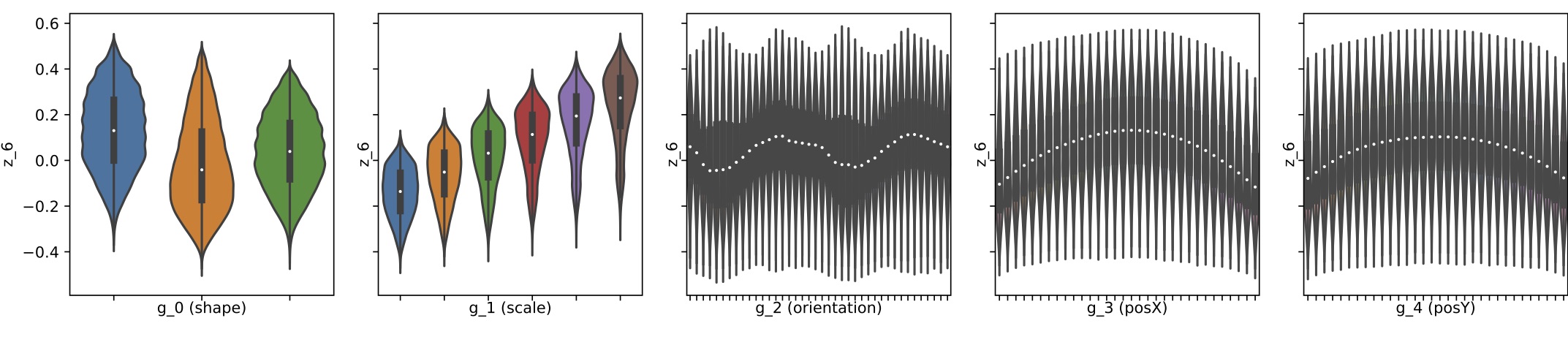}
  \caption{\textbf{Visualising Conditional Distributions:} These plots illustrate the violin plots (density estimates) of the conditional distributions $p(z_l | g_j)$ for all generative factors $G_j$ (different boxes) and for all realizations $g_j$ of $G_j$ each (x axis in each plot). The upper plot corresponds to the well disentangled and robust feature $Z_3$ of the DIP model, the lower to the disentangled (according to MI and FI) but not robust (according to $\mathrm{IRS}$) feature $Z_6$.}\label{fig:latentSpace}
\end{figure*}

A related visualization possibility to the one we propose in Section \ref{sec:visualizing} is that of simple conditioning on different generative factors (without keeping one factor fixed). This is illustrated in Figure \ref{fig:latentSpace}, where we plot the violin plots (i.e., density estimates) of $p(z_l | g_j)$ for all generative factors $G_j$ (columns) and realizations of them $g_j$ (x axis). This kind of visualization works well to discover simple dependency patterns as well as their noise levels.

\section{Visualisations of Interventional Effects}\label{sec:viseffects}
\begin{figure*}[tb!]
  \centering
    \includegraphics[height=1\textheight]{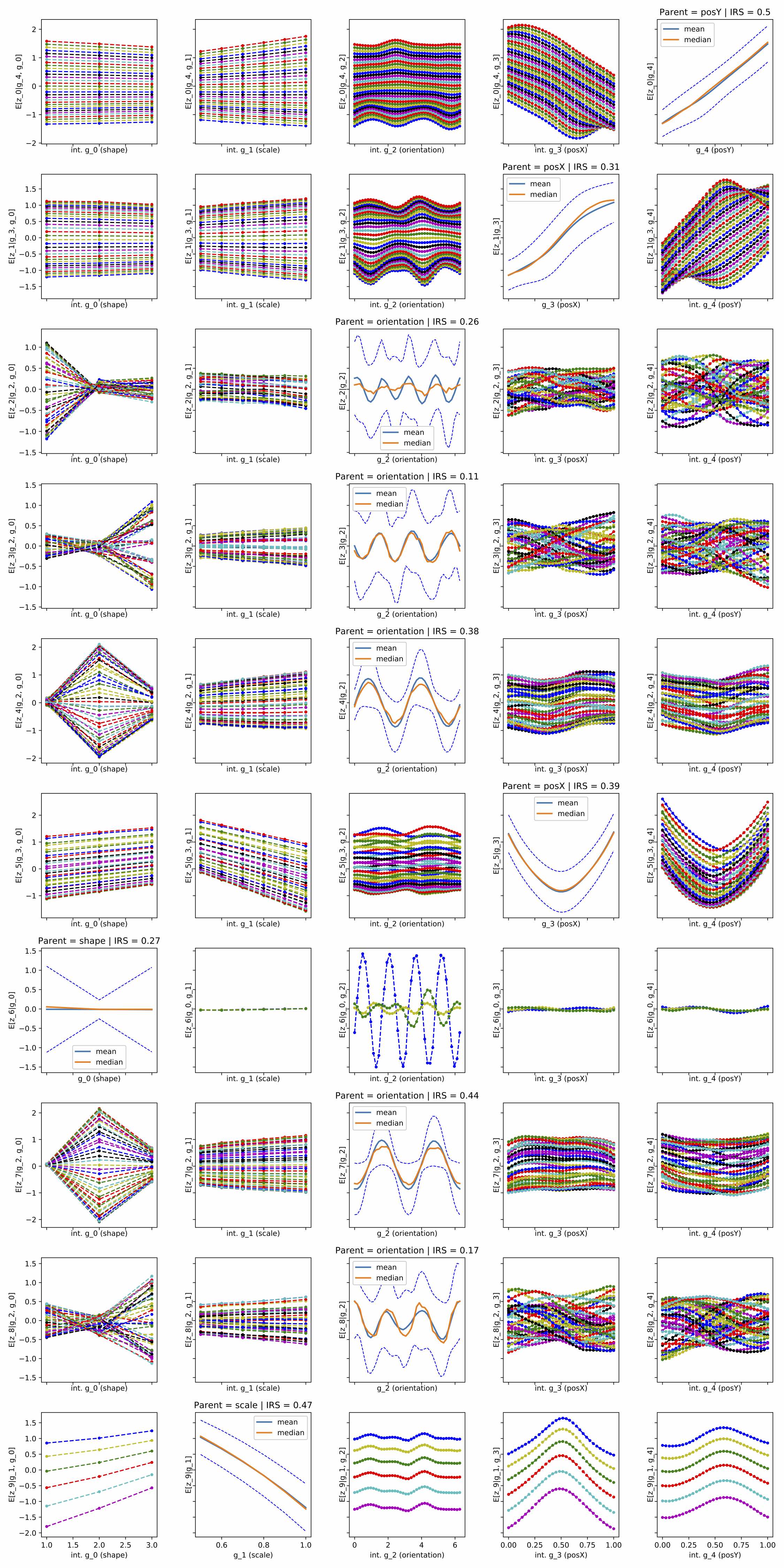}
  \caption{Visualization of interventional effects in the regular VAE model \citep{kingma2013auto}.}
  \label{fig:visVAE}
\end{figure*}

\begin{figure*}[tb!]
  \centering
    \includegraphics[height=1\textheight]{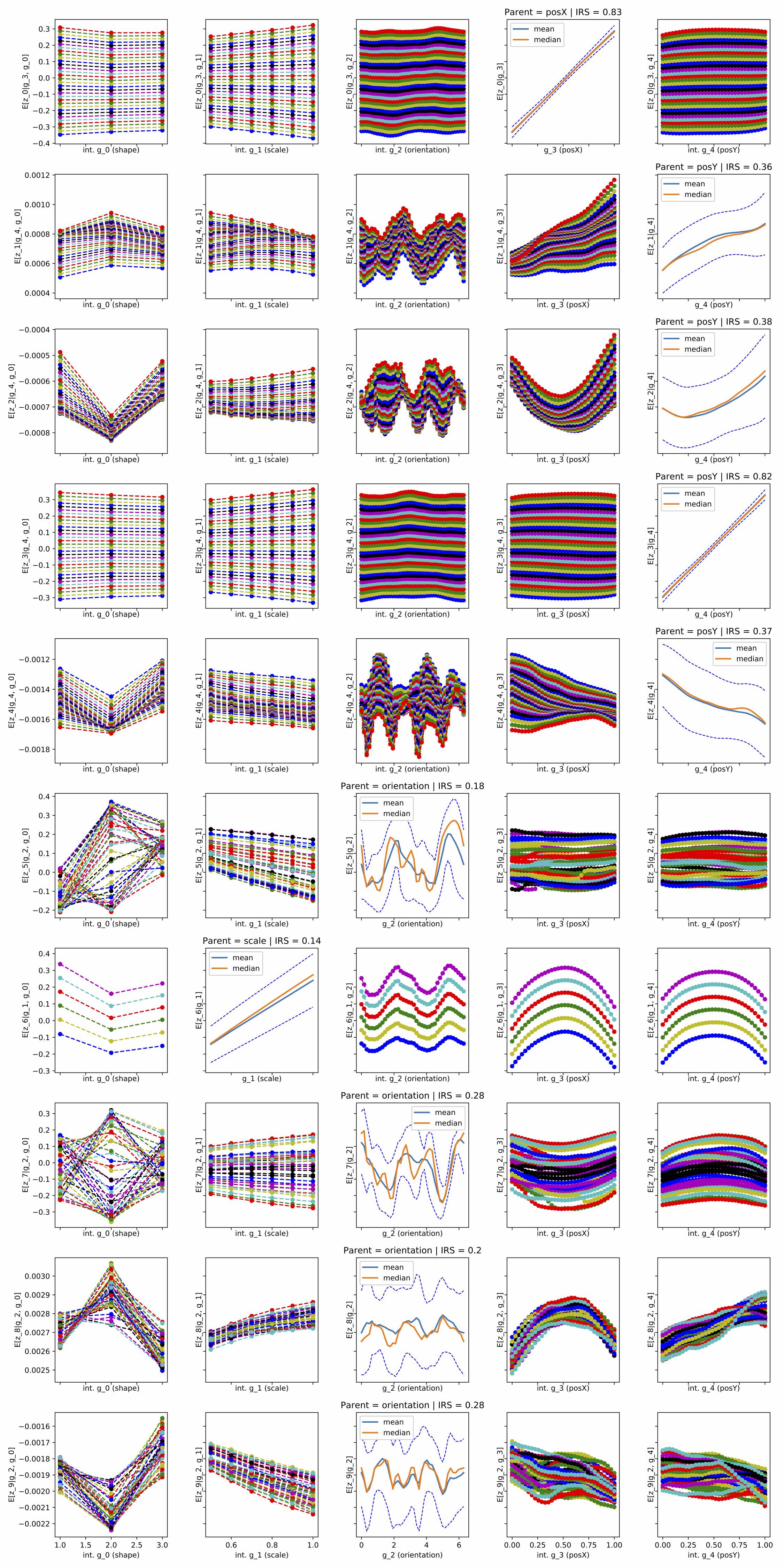}
  \caption{Visualization of interventional effects in the DIP-VAE model \citep{kumar2017variational}.}
  \label{fig:visDIP}
\end{figure*}

\begin{figure*}[tb!]
  \centering
    \includegraphics[height=1\textheight]{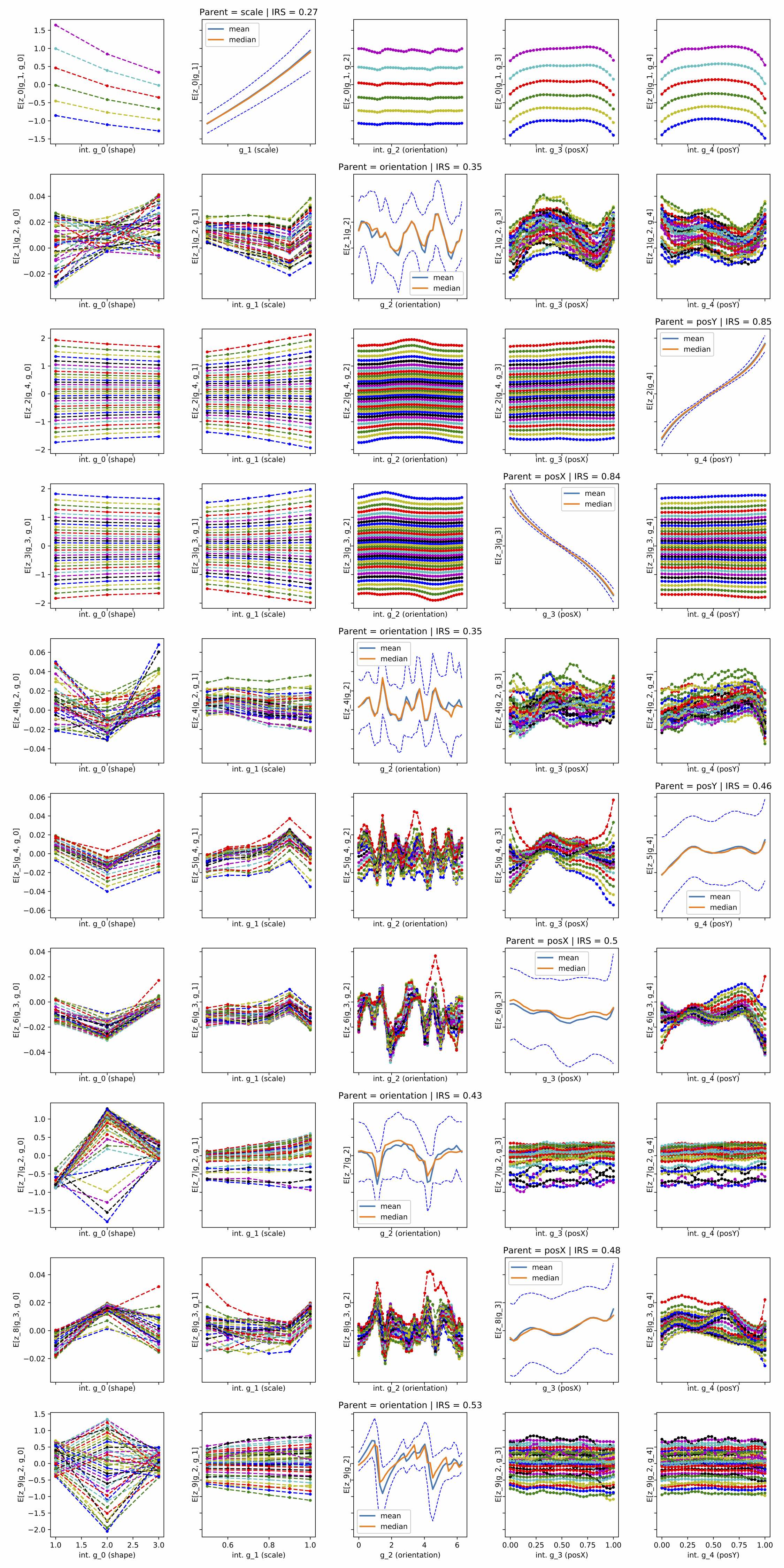}
  \caption{Visualization of interventional effects in the annealed $\beta$-VAE model \citep{higgins2016beta, burgess2018understanding}.}
  \label{fig:visannealedBeta}
\end{figure*}

\begin{figure*}[tb!]
  \centering
    \includegraphics[height=1\textheight]{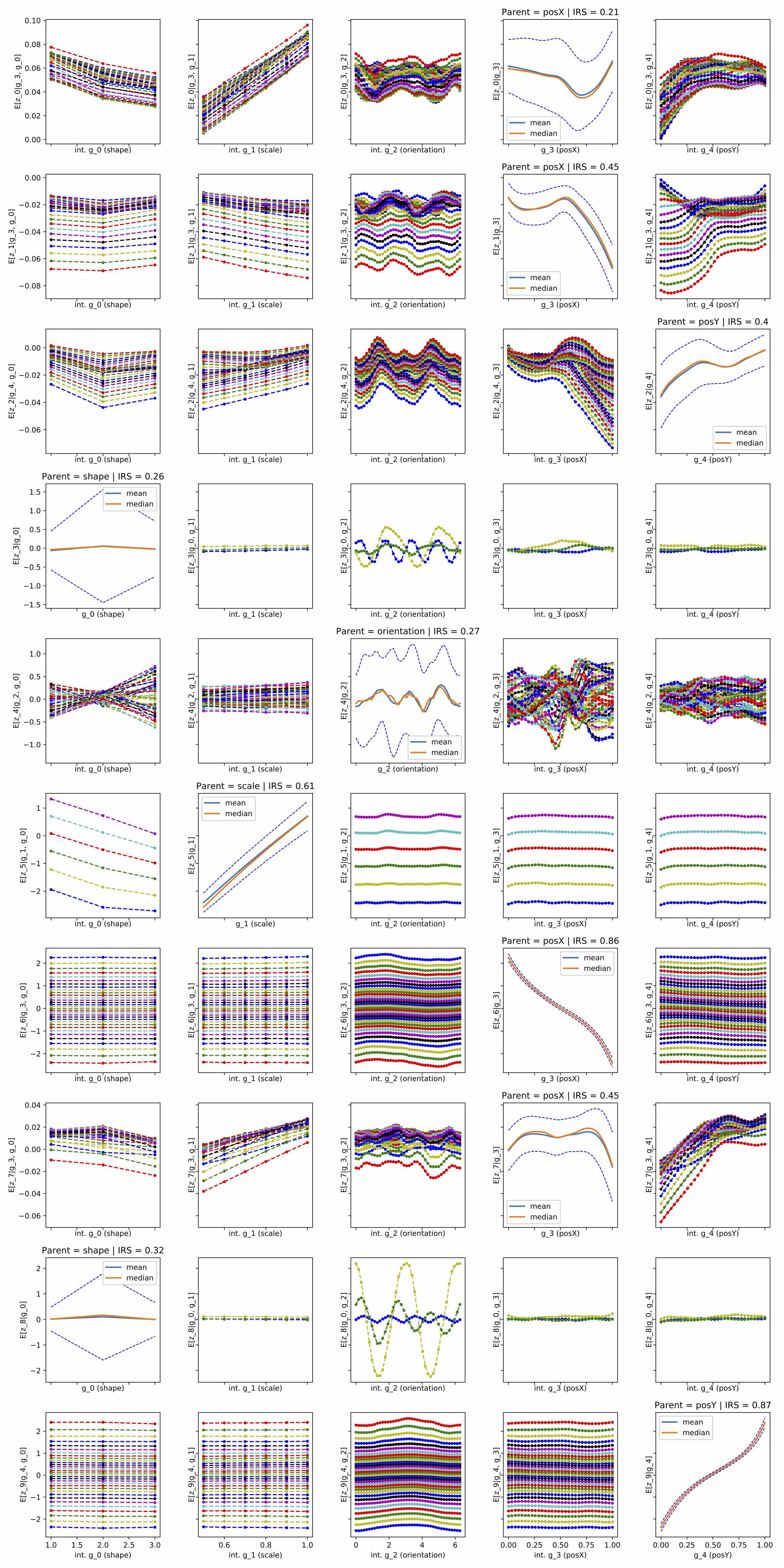}
  \caption{Visualization of interventional effects in the Factor-VAE model \citep{kim2018disentangling}.}
  \label{fig:visFactorVAE}
\end{figure*}

\begin{figure*}[tb!]
  \centering
    \includegraphics[height=1\textheight]{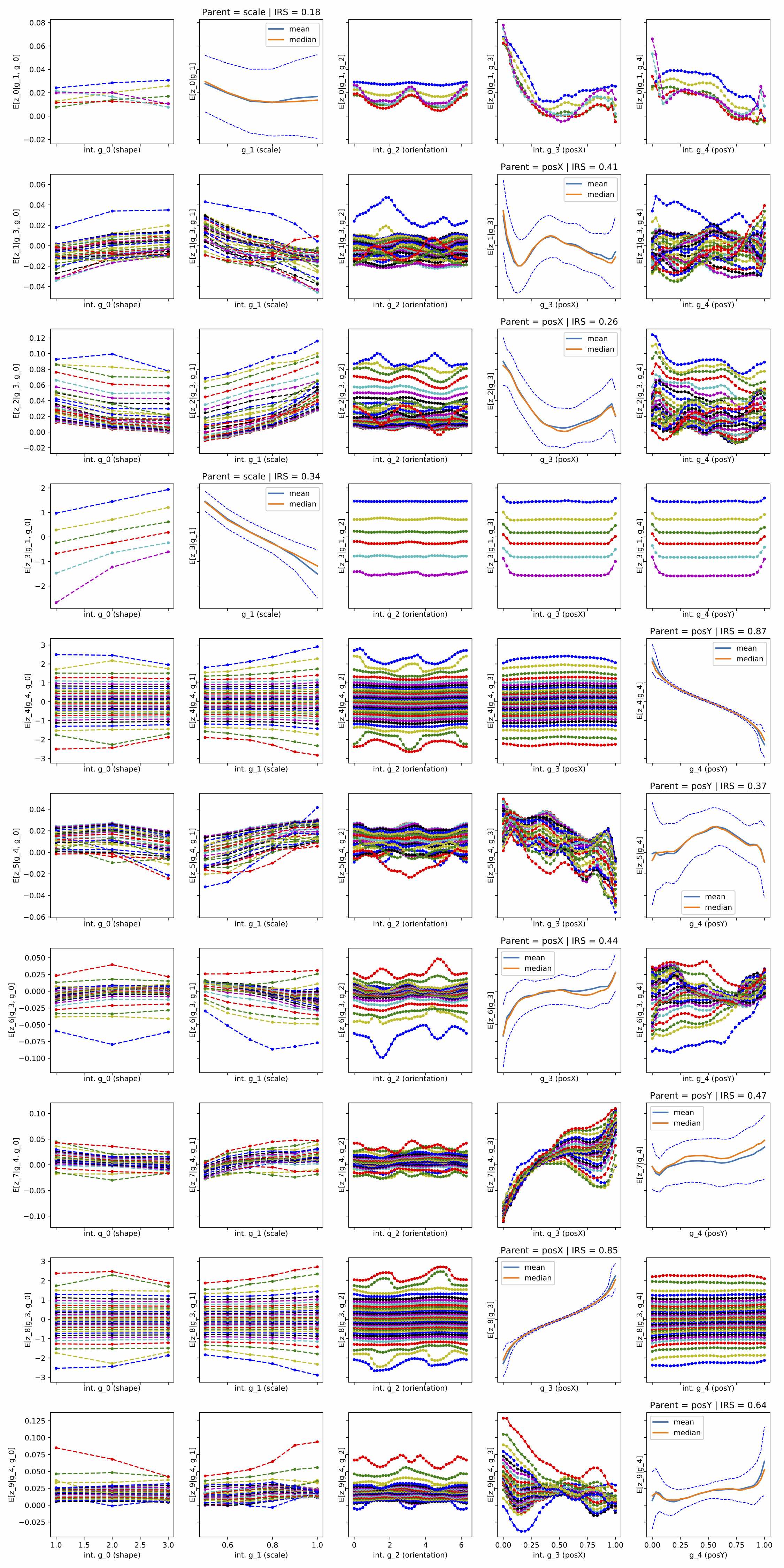}
  \caption{Visualization of interventional effects in the $\beta$-TCVAE model \citep{chen2018isolating}.}
  \label{fig:visbetatcvae}
\end{figure*}

We provide further visualizations of the full latent spaces and their dependency structure (produced by the to be made publicly available code) of a couple of models in Figures \ref{fig:visVAE}, \ref{fig:visDIP}, \ref{fig:visannealedBeta}, \ref{fig:visFactorVAE} and \ref{fig:visbetatcvae}.


\end{document}